# Geometry-aware Depth-guided Representation Learning for Structure-preserving Low-light Image Enhancement

*Fang Gao[a], Jiongkai Qin[a], Jiabao Wang[a], Jingfeng Tang[a], Ming Cheng[b,]*, Hanbo Zheng[a], Qingbao Huang[a], Cheng Wu[b,]**

*a School of Electrical Engineering, Guangxi University, Nanning, Guangxi 530004, China*

*b School of Rail Transportation, Soochow University, Suzhou, Jangsu 215500, China*

**Corresponding author*

ABSTRACT

Low-light degradation reduces image visibility and weakens structural cues that are important for visual representation and scene understanding. Existing low-light image enhancement methods mainly focus on appearance restoration, while insufficiently exploiting scene geometry to preserve structural consistency. To address this limitation, this paper proposes a Depth-guided Multi-scale Attention Network (DMSA-Net) for geometry-aware low-light image enhancement. DMSA-Net introduces depth-related structural priors into low-light representation learning through reflectance–geometry interaction. A Retinex-based decomposition module is first used to obtain illumination-invariant reflectance representations, from which depth cues are inferred to characterize scene structure under degraded illumination. A multi-scale depth-guided fusion strategy is then embedded into a hierarchical encoder–decoder architecture, where depth-aware attention adaptively integrates geometric and appearance features. Experiments on several benchmark datasets show that DMSA-Net achieves effective low-light restoration while improving structural preservation. Moreover, we construct LOL-D, a depth-augmented low-light dataset, to facilitate research on geometry-aware low-light vision.

## 1. Introduction

Low-light image enhancement (LLIE) aims to recover reliable and structurally consistent visual representations from images captured under degraded illumination conditions. In real-world computer vision applications, visual sensors frequently operate in environments with insufficient or spatially non-uniform illumination. Due to inherent imaging limitations such as restricted dynamic range, photon starvation, and reduced signal-to-noise ratio (SNR) [1], the image formation process is severely corrupted. Consequently, low-light images often suffer from luminance attenuation, contrast suppression, color distortion, and noise amplification, which degrade both photometric fidelity and structural perception.

Such degradation affects not only visual quality but also the reliability of feature representations required for image understanding. From the perspective of representation learning, adverse illumination weakens structural cues related to object boundaries, spatial layouts, and scene geometry, thereby reducing the discriminability and consistency of learned features. This limitation makes it challenging for enhancement models to preserve geometry-consistent representations during image reconstruction. As a result, downstream vision tasks, including object detection, scene understanding, and pattern recognition, may be significantly compromised when applied to degraded low-light inputs [2–5].

This work was supported by Empower Action Plan (Guangxi Key Research and Development Program) (GUIKE FN2600640198).

Fang Gao: fgao@gxu.edu.cn; Jiongkai Qin: 2412391050@st.gxu.edu.cn; Jiabao Wang: 2112401010@st.gxu.edu.cn; Jingfeng Tang: 392298627@qq.com; Ming Cheng: mcheng1@suda.edu.cn; Hanbo Zheng: hanbozheng@163.com; Qingbao Huang: qbhuang@gxu.edu.cn; Cheng Wu: cwu@suda.edu.cn.

Therefore, developing effective computational models to mitigate illumination-induced degradation without relying on hardware modifications remains a critical research challenge. In this context, LLIE can be interpreted as a representation enhancement problem, where the objective is to recover informative and geometry-consistent features from degraded inputs, thereby supporting robust perception and downstream vision tasks.

Early low-light image enhancement approaches primarily relied on histogram equalization techniques [6], [7] or models derived from Retinex theory [8-13]. These traditional methods enhance image contrast by redistributing global or local pixel intensities. However, they generally lack the capability to adapt to complex and spatially varying illumination conditions.

In recent years, deep learning–based approaches have become the dominant paradigm for low-light image enhancement. These end-to-end models learn complex nonlinear mappings from large-scale training data and have achieved remarkable performance on various benchmark datasets [14-21]. Nevertheless, when confronted with more challenging real-world scenarios involving extreme illumination conditions and complex noise characteristics, the generalization capability and robustness of existing methods remain limited. A prominent limitation is that, while improving overall brightness, many models struggle to effectively preserve fine-grained details such as edges and textures.

To address the challenges of insufficient detail preservation and limited generalization capability in low-light image enhancement, the incorporation of external prior information has emerged as an effective strategy. Existing studies have explored this direction by leveraging different forms of priors. For example, Dong et al. [22] regulated the enhancement process using generated structural cues, and Zhou et al. [23] employed perceptual priors derived from vision–language models to improve enhancement performance under low-light conditions. However, the semantic or edge-based priors adopted in these methods predominantly operate at the two-dimensional image level, making it difficult to fully characterize the spatial relationships and geometric constraints of real-world scenes, thereby limiting physical consistency and measurement stability.

Considering that depth information can be regarded as a geometric complement to RGB imaging and explicitly encodes the three-dimensional structural properties of a scene, exploring and integrating depth-related structural priors can provide effective constraints for low-light image enhancement. Such integration contributes to improved structural consistency and more reliable visual perception under complex and challenging illumination conditions.

RGB images and depth maps exhibit pronounced modal discrepancies, which lead to complementary information representations but also introduce fundamental challenges for their joint utilization. First, depth estimation is highly sensitive to illumination conditions. In low-light scenarios, depth inference models are required to simultaneously recover intrinsic reflectance properties and geometric structure from severely degraded visual signals. These two estimation tasks become strongly coupled and mutually interfering under insufficient illumination, making direct end-to-end depth estimation particularly prone to bias and instability. Second, once a depth map is obtained, effectively integrating depth information as a geometric prior with the color and texture cues of RGB images across different network hierarchies remains nontrivial. Simple fusion strategies, such as feature concatenation or element-wise addition, are inadequate to accommodate the varying modal contributions across diverse scenes and feature scales. Inappropriate fusion may result in cross-modal interference or even conflict, where noise in depth measurements contaminates RGB features, or low-light artifacts in RGB images distort geometric representations, thereby limiting overall model performance. Consequently, the effective incorporation of depth priors into the low-light image enhancement process constitutes a critical and challenging task.

To address the aforementioned challenges, this paper proposes a Depth-aware Multi-scale Attentional Network (DMSA-Net) for low-light image enhancement. The proposed network adopts an end-to-end architecture that explicitly incorporates depth information as a geometric prior and leverages a multi-scale attention mechanism for feature fusion, enabling more accurate and perceptually natural enhancement results. The overall framework consists of two core components: (1) a front-end depth estimation module, and (2) a multi-scale attentional encoder–decoder fusion module.

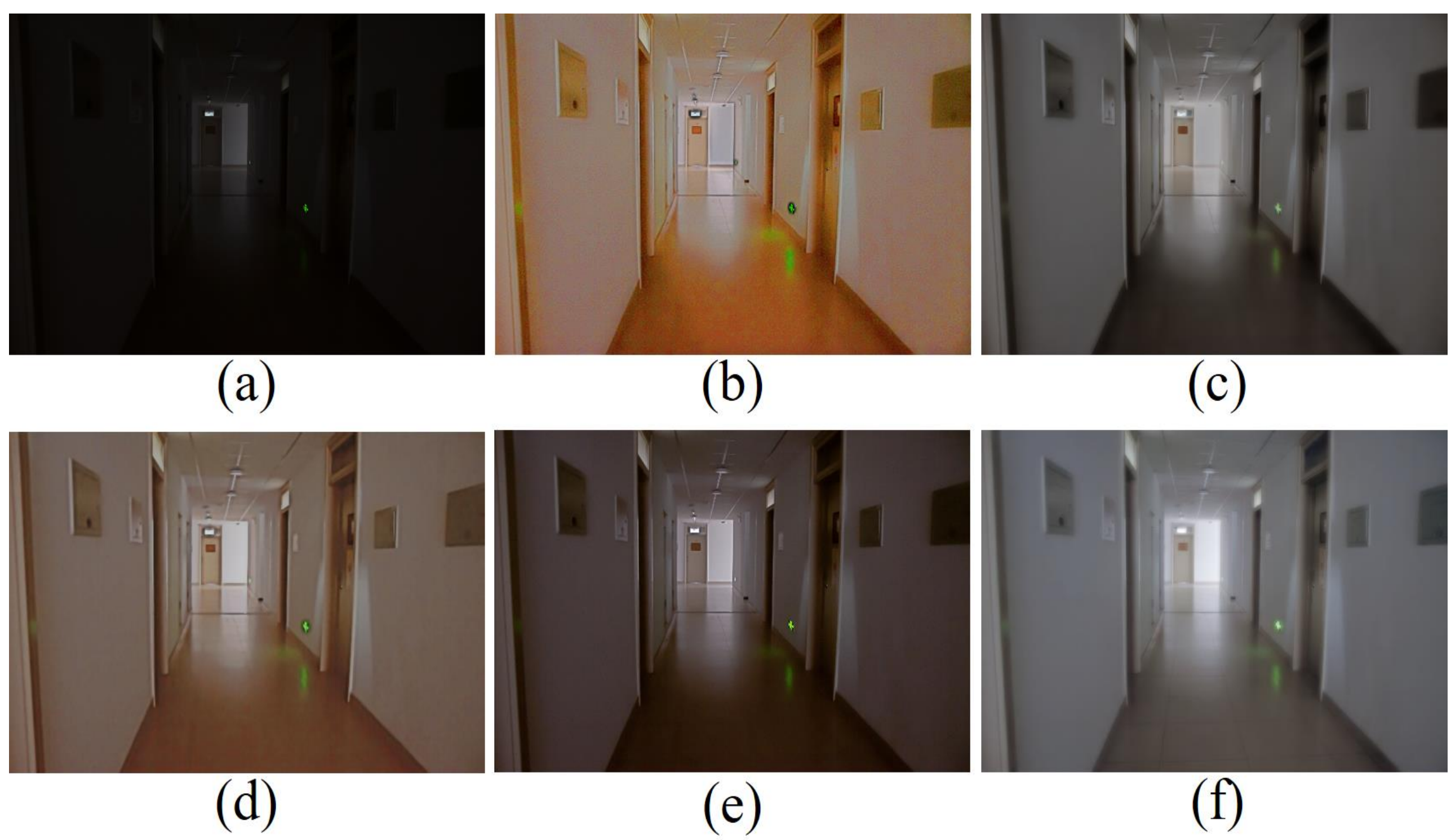

Fig. 1. (a) Input image taken in low-light environments. (b)-(e) Image enhancement through other advanced existing methods (Retinex-Net, KinD, PairLIE, and SCI, respectively). (f) Image enhancement using our method (DMSA-Net)

The front-end depth estimation module is built upon Retinex decomposition and advanced visual foundation models to extract robust geometric priors from input low-light images. Specifically, this module first applies a pre-decomposition network to decouple the low-light image into an illumination component (L) and a reflectance component (R). The reflectance component encodes the intrinsic texture and structural information of the scene and remains invariant to illumination changes. This reflectance component is then fed into a state-of-the-art monocular depth estimation model, effectively mitigating the influence of lighting variations and enabling stable and accurate depth inference.

After obtaining a reliable depth prior, the next critical step is to efficiently and adaptively integrate it as a complementary modality into the low-light image restoration features. Inspired by Lin et al. [24], we perform depth feature fusion exclusively within the encoder. This design choice is motivated by the observation that encoder features are more generalizable and align well with depth priors, whereas decoder features are specialized for target-domain reconstruction, and fusing depth at the decoder stage offers limited benefits and may introduce noise due to the semantic gap.

Accordingly, we develop a Multi-scale Depth Fusion (MDF) block at each level of the encoder. Each MDF block consists of a three-stage progressive feature processing pipeline: First, leveraging the parallel multi-branch structure of the Multi-branch Attention Module (MbAM), the low-light input image is encoded to obtain robust feature representations with rich receptive fields. On this basis, we innovatively introduce the Depth-aware Attentional Feature Fusion (DAAF) module, which employs a cross-attention mechanism to adaptively inject the geometric priors embedded in the depth map into the image feature space, thereby providing precise geometric constraints for the low-light image enhancement process. Additionally, the module incorporates Residual Dense Blocks (RDB) to efficiently recover texture details lost under low-light conditions, further improving the perceptual quality of the output images.

We conducted extensive experiments on multiple low-light image datasets. The experimental results demonstrate that our method achieves superior restoration performance. Fig. 1 visually compares our method with other cutting-edge methods. In summary, our contributions are fourfold:

1) We propose a decoupled depth estimation framework that introduces a front-end decomposition to separate geometric depth estimation from illumination interference removal, enabling robust intrinsic scene representations for low-light enhancement.

2) We introduce a Multi-scale Depth Fusion (MDF) block that combines self-attention for global context modeling and cross-attention for image–depth correlation learning, ensuring accurate and comprehensive multimodal feature fusion.

3) We design a plug-and-play depth-aware attentional feature fusion (DAAF) module that adaptively fuses image and depth features via cross-modal attention, effectively exploiting geometric structure priors for enhanced low-light restoration.

4) We extend the LOL dataset to LOL-D using our proposed low-light depth estimation module, which includes LOLv1-D and LOLv2-D. The dataset provides corresponding depth maps for all image pairs, thereby supporting research on depth-aware low-light image enhancement. It will be publicly available at: https://github.com/toppyuser/LOL-D.

## 2. Related work

### *2.1. Traditional enhancement methods*

Traditional low-light image enhancement methods primarily rely on manually designed heuristic models and enhancement strategies. Histogram equalization (HE)–based approaches [25] redistribute pixel intensity values to expand the dynamic range [7], [26]-[28], while gamma correction [29] adjusts overall brightness through nonlinear mapping functions [30]. Another important line of work is based on Retinex theory [8], which decomposes an image into reflectance and illumination components. By enhancing the illumination component and regularizing the reflectance, these methods can improve brightness while effectively suppressing noise, achieving notable results in exposure correction [10], [11], [13], [31], [32], [33]. Such traditional approaches perform well under relatively uniform illumination conditions. However, in real-world low-light scenarios, lighting is often highly dynamic and heterogeneous, significantly limiting their effectiveness. Moreover, designing new hand-crafted priors for complex scenes is both challenging and labor-intensive. These limitations have collectively driven the development of data-driven deep learning approaches in this domain.

### *2.2. Deep learning-based enhancement method*

With the widespread adoption of deep learning in this domain, various methodological paradigms have emerged [34]-[39]. Supervised learning approaches leverage paired “low-light/normal-light” image data to train enhancement models. Lore et al. [40] employed stacked sparse denoising autoencoders to identify signal features from low-light images and adaptively brighten them in high dynamic range images. Wei et al. [41] developed the first real paired dataset, LOL-v1, and trained an end-to-end network in a fully supervised manner. Lv et al. [42] proposed a multi-branch low-light enhancement network that extracted rich features at different hierarchical levels, employed multiple sub-networks to perform enhancement tasks, and fused branch-level features to produce the final output. Feng et al. [43] introduced a dual-branch structure with progressive block fusion to effectively mitigate over-smoothing, noise, and block artifacts in low-light enhancement. Zhang et al. [44] decoupled images into grayscale and color histogram components and leveraged an innovative pyramid color embedding module to address color distortion, achieving superior color consistency across multiple datasets.

To reduce reliance on hard-to-obtain paired data, unsupervised learning paradigms have also been explored. Fu et al. [45] constrained reflectance consistency across low-light image pairs to adaptively learn priors from data, achieving comparable enhancement with simpler network architectures. Wu et al. [46] combined Retinex theory with CycleGAN to perform joint correction of illumination, noise, and color distortions using unpaired data, producing satisfactory results. Ma et al. [47] introduced a weight-sharing self-calibrated module that enabled cascaded training and single-stage inference, improving both efficiency and scene adaptability while maintaining enhancement performance. However, deep learning networks alone often lack explicit physical guidance or reliable stability constraints during the enhancement process. Consequently, their generalization and robustness remain limited when faced with the complex and variable conditions of real-world low-light imaging scenarios.

### *2.3. Methods with Prior Knowledge*

A series of studies have attempted to incorporate various forms of prior knowledge to construct more interpretable enhancement models. Yang et al. [48] designed an end-to-end network guided by signal priors to separate layers, combined with a data-driven mapping mechanism, where layer-specific constraints enable efficient enhancement of single low-light images. Jin et al. [49] leveraged deep structural features and depth inconsistency priors to address the fusion challenges caused by noise and structural mismatches under low-light conditions. Yang et al. [50] employed implicit neural representations with semantic-guided supervision and dual-loop constraints to achieve robust unsupervised enhancement for unpredictable degradations, generating perceptually favorable results. Xu et al. [51] and Dong et al. [22] utilized structural information to guide the low-light enhancement process, improving both detail recovery and visual quality.

However, these priors are predominantly extracted and fused in a two-dimensional context, neglecting the three-dimensional information inherent in the scene. Recently, Lin et al. [24] introduced an innovative approach in low-light enhancement by designing hierarchical depth feature extraction and fusion modules, incorporating depth priors as core geometric information into the enhancement pipeline for the first time. This work provides a novel paradigm for multimodal information-driven optimization, demonstrating that explicit geometric priors can significantly improve low-light enhancement outcomes.

## 3. Proposed method

This section systematically presents the theoretical framework of the Depth-aware Multi-scale Attentional Network. First, in Section 3.1, we discuss the motivation for incorporating depth information and provide an overview of the overall network architecture. Next, in Section 3.2, we detail how depth information is robustly extracted from low-light images through a front-end task based on the Retinex framework. Subsequently, Section 3.3 focuses on a thorough analysis of the design of the proposed MDF block. Finally, Section 3.4 specifies the model training methodology and defines the roles of the various loss functions employed.

### *3.1. Motivation*

Traditional low-light image enhancement methods and their early deep learning variants generally operate solely within the two-dimensional image space [7], [8], [10], [36], [44], [48], [50]. While these approaches have achieved moderate success, their fundamental limitation lies in treating images as planar signal collections without three-dimensional scene understanding. Consequently, two-dimensional models lack awareness of the geometric structure of the scene and are unable to adaptively enforce enhancement constraints based on surface continuity or object boundary variations.

In contrast, depth maps provide direct information about scene layout and relative distances between objects, offering critical spatial structure priors that can guide the enhancement process more effectively.

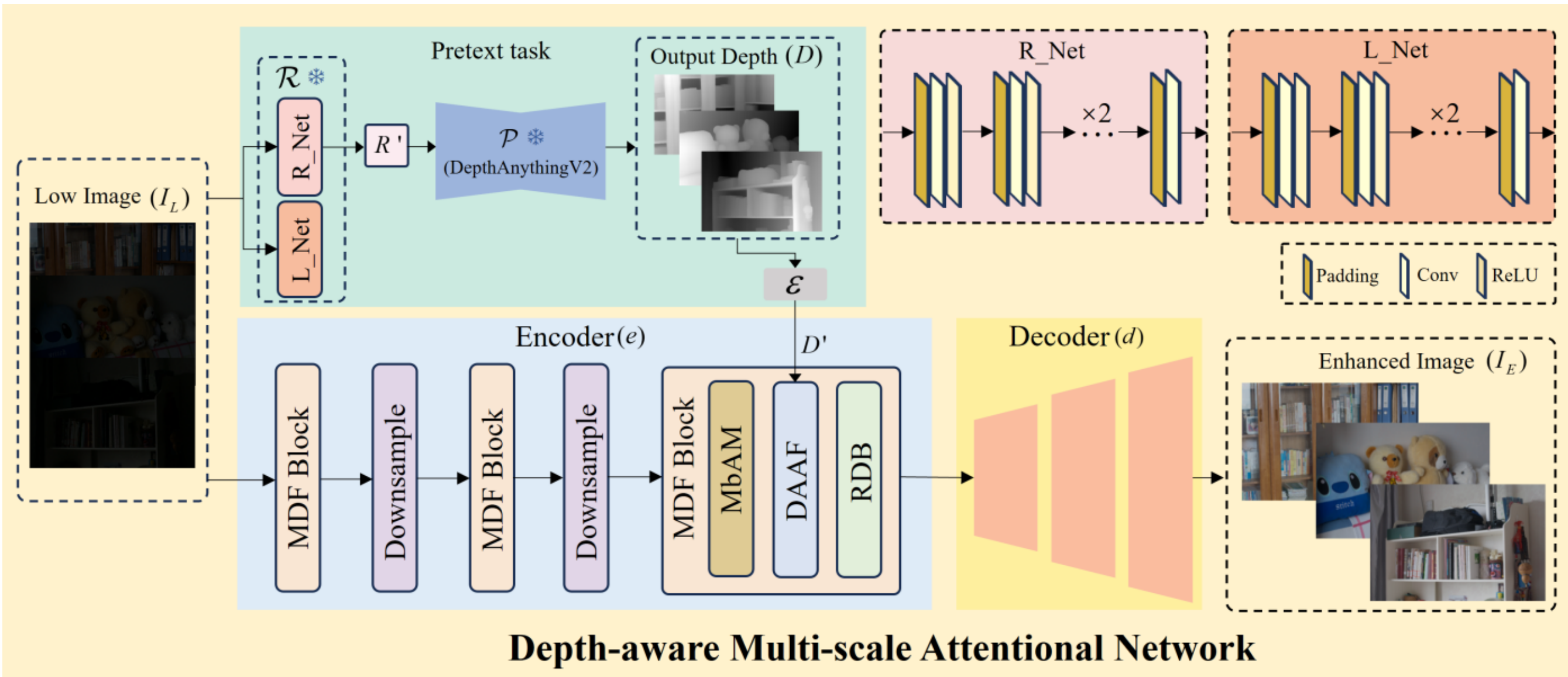

Fig. 2. Overview of the proposed Depth-aware Multi-scale Attentional Network (DMSA-Net) for low-light image enhancement. The network consists of a front-end Retinex-based decomposition module that separates the input image into reflectance and illumination components, followed by robust depth estimation from the illumination-invariant reflectance. Extracted depth maps are integrated with the RGB features produced by the Multi-branch Attention Module (MbAM) through the Depth-Aware Attentional Feature Fusion (DAAF) module within the encoder–decoder framework, enabling structure-aware enhancement. Residual dense blocks (RDB) are embedded to preserve fine-grained texture details. The overall design ensures adaptive utilization of geometric priors for improved brightness, contrast, and geometric consistency under challenging low-light conditions.

Specifically, depth maps explicitly delineate the contours and spatial positions of different objects within a scene. Leveraging this information, the network can learn to distinguish flat regions from structural edges, thereby enabling smooth suppression in uniform areas to avoid noise amplification while simultaneously preserving object boundaries to maintain structural integrity. This is equivalent to endowing the enhancement model with a form of 'spatially adaptive' capability, allowing its enhancement strategy to align with the geometric structure of the scene.

Consequently, depth information is required to serve as a guide for fusion weighting. By analyzing depth features, the model can dynamically determine how RGB appearance features and depth geometric features should contribute to the final reconstruction at different scales. For instance, in regions with abrupt depth changes, the model should rely more heavily on depth features to preserve geometric structure, whereas in homogeneous regions with smooth depth, RGB features can be prioritized for texture recovery. This depth-guided adaptive fusion mechanism significantly enhances both the efficiency and effectiveness of feature utilization.

Our overall framework is illustrated in Fig. 2. The model is designed to leverage illumination-invariant depth information as a geometric prior to guide effective low-light image enhancement. Its core process can be formulated as a cascaded learning framework, comprising a preliminary depth estimation task followed by a multi-scale attention-based encoder-decoder fusion module.

### *3.2. Extraction of depth information*

Reliable depth information, serving as a geometric prior, is crucial for guiding low-light image enhancement models to achieve spatially adaptive restoration. However, under extremely low illumination, directly estimating accurate monocular depth maps from low-light images is highly challenging due to the drastic drop in signal-to-noise ratio and color distortion. State-of-the-art monocular depth estimation models, such as DepthAnythingV2 [52], UniDepth [53], and Depthpro [54], perform well on normally illuminated images but exhibit significant performance degradation in low-light conditions, as they do not explicitly model the domain shift caused by severe illumination variations.

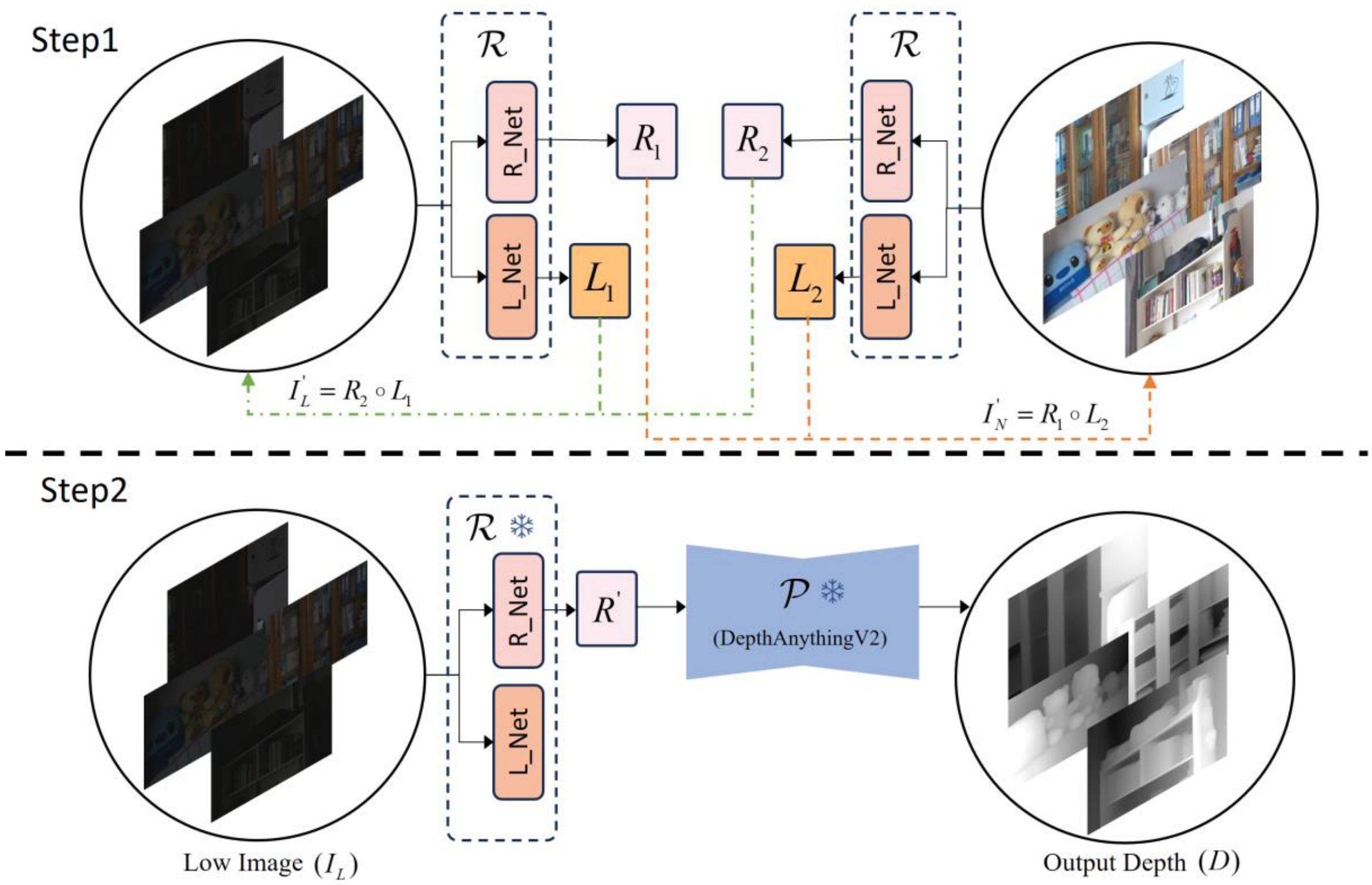

Fig. 3. An overview of the pretext task for extracting depth information. The task consists of two steps: (1) A Retinex-based decomposition network $\mathcal{R}$ is trained to obtain the reflectance component $R$and the illumination component $L$; (2) The network $\mathcal{R}$ parameters are frozen, and depth information with physical invariance is extracted from the $R$ component using the DepthAnythingV2 network, ultimately generating the depth map.

As described by the Retinex theory, the image $I$ can be decomposed into intrinsic reflectance and extrinsic illumination components:

$$I = R \circ L \tag{1}$$

The reflectance component $R$ encodes the inherent visual properties of objects and is invariant to changes in lighting. Based on this theory, we design a two-stage pretext task, as illustrated in Fig. 3. First, a Retinex decomposition model $\mathcal{R}$ is constructed to decompose the input low-light image into the reflectance $R$ and illumination components $L$, and the model is trained with a simple pre-training procedure. Next, the parameters of $\mathcal{R}$ are fixed, and during deployment, the low-light image is fed into the model to extract the illumination-invariant reflectance $R'$. Finally, a pre-trained monocular depth estimation model $\mathcal{P}$ is applied to $R'$ to estimate depth, thereby obtaining a robust, illumination-invariant depth prior $D$.

### *3.2.1. Decomposition model in pretext task*

The Retinex decomposition network $\mathcal{R}$ in the pretext task serves as the fundamental module for effective component disentanglement. It consists of two structurally symmetric but weight-independent subnetworks, $L_Net$ and $R_Net$, denoted as $\mathcal{F}_R$ and $\mathcal{F}_L$, which are responsible for extracting the illumination component $L$ and the reflectance component $R$ from the input image, respectively. As shown in the top-right of Fig. 2, inspired by ref. [45], each subnetwork is composed of a stack of simple convolutional modules to progressively extract features. Given a low-light image $I_L$ and a normal-light image $I_N$, the decomposition process can be formulated as follows:

$$\begin{aligned} R_1, L_1 &= \mathcal{F}_R(I_L), \mathcal{F}_L(I_L) \\ R_2, L_2 &= \mathcal{F}_R(I_N), \mathcal{F}_L(I_N) \end{aligned} \tag{2}$$

Here $R_1$ and $L_1$ denote the decomposition results of the low-light image $I_L$, while $R_2$ and $L_2$ denote those of the normal-light image $I_N$. The input image should also be reconstructable from the decomposed components, i.e.:

$$\begin{aligned} I_L &\approx R_1 \circ L_1 \\ I_N &\approx R_2 \circ L_2 \end{aligned} \tag{3}$$

The reflectance component $R$ is defined as the intrinsic physical property of the scene, describing the surface reflectance of objects to light of different wavelengths, independent of the illumination conditions during imaging. Therefore, for two images captured of the same scene under different lighting conditions, their reflectance components should ideally be identical, i.e., $R_1 = R_2$. This physical prior forms the basis of our training process.

#### *3.2.2. Training strategy of decomposition model*

Our model is trained on a dataset composed of paired low-light images $I_L$ and normal-light images $I_N$. The training process adopts a self-supervised strategy based on cycle consistency, with the core objective of ensuring that the decomposed reflectance component remains illumination-invariant. The specific procedure corresponds to Step 1 in Fig. 3.

During training, the image pairs are fed into the decomposition network $\mathcal{R}$ to obtain their respective reflectance components $R_1, R_2$ and illumination components $L_1, L_2$. Subsequently, the core cycle reconstruction operation is performed by swapping the reflectance components and performing element-wise multiplication with the original illumination components:

$$\begin{aligned} \hat{I}_L &= R_2 \circ L_1 \\ \hat{I}_N &= R_1 \circ L_2 \end{aligned} \tag{4}$$

The physical essence of our cycle reconstruction lies in a strong constraint: it enforces that the reflectance components $R_1$ and $R_2$ learned by the decomposition network must closely approximate the true shared reflectance $R$. When used for cross-reconstruction, they should accurately restore the original input images. This constraint compels the network to learn illumination-invariant, highly consistent intrinsic reflectance features of the scene, providing more robust and accurate inputs for downstream tasks such as depth estimation, while effectively enhancing the generalization capability and physical interpretability of the decomposition model.

The loss function of the Retinex decomposition model in the pretext task consists of three components: the decomposition loss, the reflectance consistency loss, and the cycle reconstruction loss, which together guide the model to learn decomposition representations that conform to physical priors. The decomposition loss ensures that the decomposed components can accurately reconstruct the original image while imposing a smoothness constraint on the illumination component:

$$L_d = \|R \circ L - I\|_2^2 \tag{5}$$

The reflectance consistency loss enforces that the reflectance components of the same scene remain consistent under different lighting conditions:

$$L_R = \|R_1 - R_2\|_2^2 \tag{6}$$

The cycle reconstruction loss requires that the images reconstructed after swapping the reflectance components, $\hat{I}_L$ and $\hat{I}_N$, respectively approximate the original normal-light image $I_N$ and low-light image $I_L$ in content:

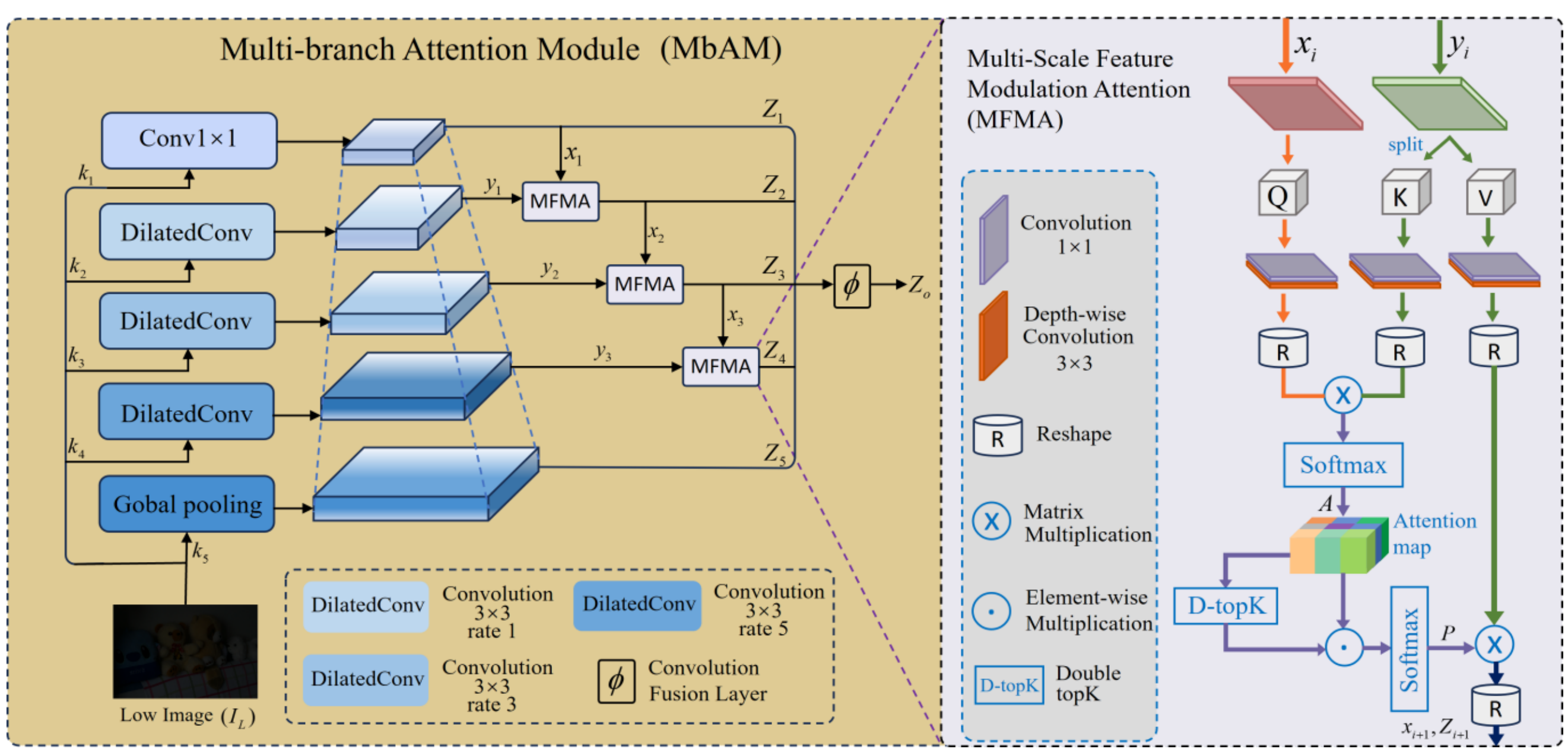

Fig. 4. Architecture of the proposed Multi-branch Attention Module (MbAM) with the embedded Multi-scale Feature Modulation Attention (MFMA). MbAM captures multi-scale context through parallel convolutions, while MFMA enhances structure-aware features via sparse cross-attention with dual Top-K selection under low-light conditions.

$$L_c = \left\|\hat{I}_N - I_N\right\|_1 + \left\|\hat{I}_L - I_L\right\|_1 \tag{7}$$

The final loss function is defined as:

$$L_{rt} = L_d + \chi_1 L_c + \chi_2 L_R \tag{8}$$

Here $\chi_1$ is set to 0.1 and $\chi_2$ is set to 0.5. By optimizing this objective, the model is able to decompose low-light images and provide illumination-invariant reflectance components for subsequent depth estimation.

### *3.2.3. Depth estimation*

After the training of the Retinex-based preliminary decomposition network $\mathcal{R}$ is completed, the depth extraction process is conducted entirely based on its output reflectance component. Specifically, the parameters of the decomposition network $\mathcal{R}$ are first frozen, and the low-light image $I_L$ is fed into the network to extract the reflectance component $R'$, which encodes the intrinsic physical properties of the scene. Subsequently, a pre-trained DepthAnythingV2 model $\mathcal{P}$ is employed as the depth estimator to predict the depth map $D$ from the reflectance component $R'$. The above procedure corresponds to Step 2 in Fig. 3. The above process can be described as follows:

$$D = \mathcal{P}(\mathcal{R}(I_L)) \tag{9}$$

This design effectively isolates depth estimation from low-light interference, yielding more reliable structural priors. Due to space limitations in the main text, additional results are provided in the supplementary materials, where depth maps predicted from low-light images are compared with those obtained under normal illumination conditions. The results demonstrate a high degree of consistency between the two, validating the robustness of the proposed approach.

### *3.3. The Multi-scale Depth Fusion (MDF) block*

In the previous section, we obtained robust depth information that is independent of illumination through the pretext task. This section elaborates on how this geometric prior is effectively integrated into the image enhancement process. As illustrated in Fig. 2, the depth information is injected into the encoder to guide spatially adaptive feature restoration.

Although the depth map explicitly captures the geometric contours and spatial layout of the scene, the complete information required to recover a normal-light image extends far beyond geometry alone, encompassing rich texture details, illumination intensity distributions, and global contextual semantics. To jointly exploit the depth prior and diverse visual features from RGB images, we introduce a core multi-scale depth fusion module within the encoder. This module primarily consists of a multi-branch attention module, a depth-aware attention fusion module, and a residual dense block (RDB) module. The RDB [55] module is solely used to process features for enhancing local representations.

#### *3.3.1. The Multi-branch Attention Module (MbAM)*

The core design objective of the MbAM module is to fully exploit the low-light RGB input to extract multi-scale appearance features enriched with contextual information, and to employ cascaded multi-scale feature modulation attention modules to achieve cross-level feature selection and fusion. This design enables the collaborative modeling of local details, structural information, and global illumination priors. The overall architecture of the MbAM is illustrated in Fig. 4.

Given a low-light input image $I_L \in \mathbb{R}^{H\times W\times C}$, the MbAM module first constructs multiple parallel feature branches to extract multi-scale feature representations with different receptive fields. Specifically, the module consists of a 1×1 convolution branch, three 3×3 dilated convolution branches with different dilation rates($r = 1, 3, 5$), and a global average pooling branch. The parallel branches jointly capture complementary information.

Direct concatenation of multi-scale features alone is insufficient to fully exploit the intrinsic correlations among different scales. To address this issue, MbAM introduces a Multi-scale Feature Modulation Attention (MFMA) module between adjacent scale features to enable cross-scale adaptive feature enhancement. Specifically, the feature $Z_1$ extracted from the $k_1$branch is regarded as the shallowest-scale representation, which contains abundant local detail information. Since shallow features do not require guidance from other scales, it is directly used as the initial state of the recursive process, i.e., $x_1 = Z_1$.

The MFMA module adopts a modulation strategy in which lower-level features guide higher-level features. Let $x_i$ and $y_i$ denote two adjacent-scale features. MFMA utilizes the lower-level feature $x_i$ as guidance to perform attention-based modulation on the corresponding higher-level feature $y_i$. Taking the first MFMA module as an example, the output $x_1$ from the preceding $k_1$ branch is used as the query input of the MFMA module in the $k_2$ branch, where it interacts with the feature $y_1$ extracted at the current scale. Subsequently, the output of each MFMA module is fed as the query input to the next MFMA module, thereby enabling recursive cross-scale feature modulation. This process can be formulated as:

$$Z_{i+1} = x_{i+1} = MFMA(x_i, y_i)(i = 1, 2, 3) \tag{10}$$

It is worth noting that feature modulation attention is introduced only in the $k_2$, $k_3$, and $k_4$ branches. For the feature $Z_5$ produced by the global pooling branch, the complete scene-level contextual information has already been aggregated through global pooling; therefore, no additional cross-scale modulation is required.

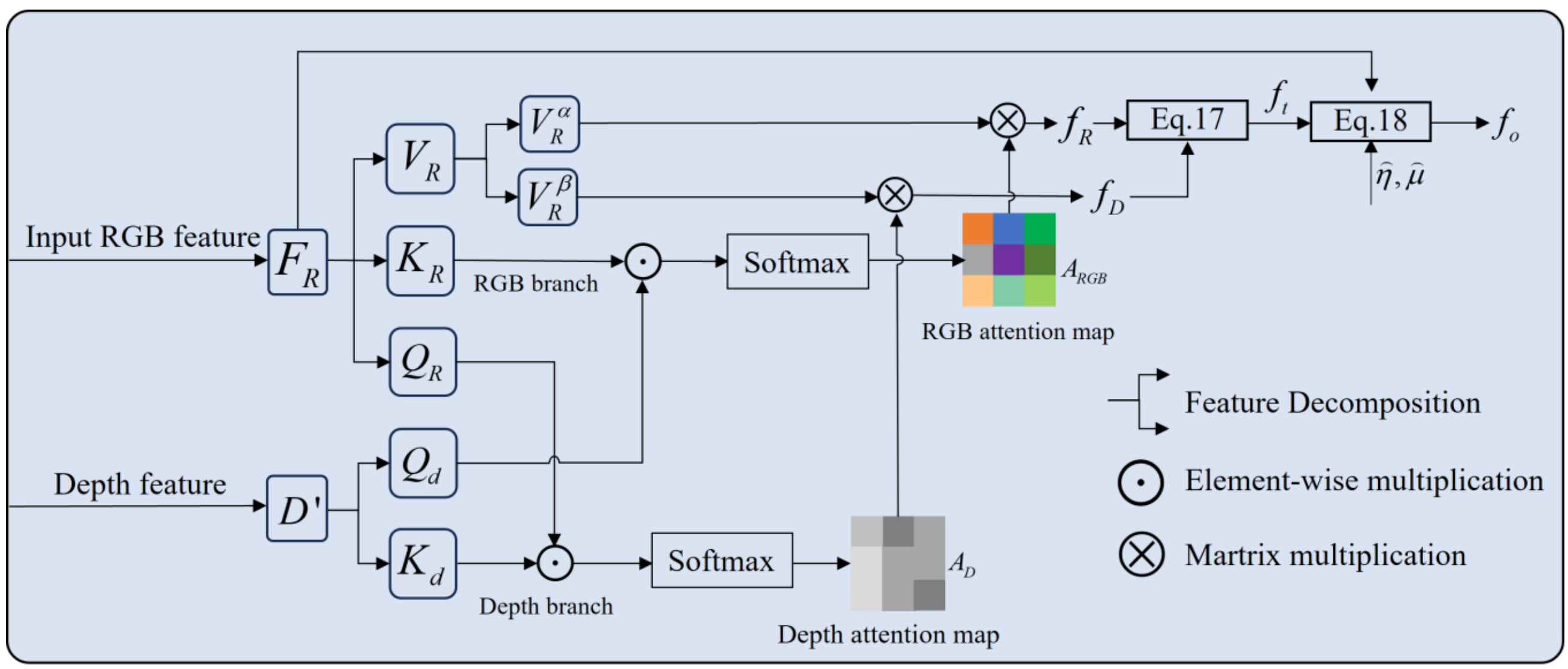

Fig. 5. Overview of the proposed Depth-Aware Attention Fusion (DAAF) mechanism. The module jointly models RGB appearance features and depth features through parallel attention branches, and adaptively fuses them via attention-guided feature decomposition and modulation to enhance structural consistency and illumination robustness in low-light image enhancement.

Within the MFMA module, representations from different feature hierarchies or semantic spaces are taken as inputs. By constructing sparse attention maps, the module explicitly models the dependencies between the query features and the key–value features, thereby enabling selective enhancement of critical information while effectively suppressing redundant responses.

Specifically, given the input features $x_i$ and $y_i$, MFMA first projects $x_i$ into the query feature $Q$, while $y_i$ is partitioned and separately projected into the key feature $K$ and the value feature $V$. Subsequently, the correlation between the query feature $Q$ and the key feature $K$ is computed via scaled dot-product attention, followed by Softmax normalization to obtain the initial attention map $A$:

$$A = \mathrm{Softmax}(\frac{QK^T}{\sqrt{d}}) \tag{11}$$

Here $d$ denotes the embedding dimension.

On this basis, MFMA introduces a D-TopK [56] sparsity constraint to perform saliency-based filtering on the initial attention map. The resulting sparse attention matrix $P$ can be expressed as:

$$P = \mathrm{Softmax}(A \odot [\mathrm{TopK}_{\mathrm{t1}}(A) + \mathrm{TopK}_{\mathrm{t2}}(A)]) \tag{12}$$

Here the TopK operation with parameters $t1 = 3$ and $t2 = 5$ is used to select the most salient multi-scale responses along the spatial dimension. The subsequent Softmax operation is applied to re-normalize the sparse attention.

After obtaining the sparse attention matrix $P$, MFMA modulates the value feature $V$ using this attention to achieve selective cross-feature interaction. The output is defined as:

$$Z_{i+1} = x_{i+1} = P \otimes V \tag{13}$$

Here $Z_{i+1}$ denotes the output feature representation of the $(i+1)$-th MFMA branch, and $x_{i+1}$ represents the query feature passed to the next stage. $\otimes$ denotes matrix multiplication.

Finally, the modulated outputs from all levels of the MbAM module are concatenated and fused through convolution to form the final output of the module:

$$Z_o = \phi(Z_1 \oplus Z_2 \oplus Z_3 \oplus Z_4 \oplus Z_5) \quad (14)$$

Here $\oplus$ denotes feature concatenation, and $\phi$ represents the fusion convolution layer. This operation effectively integrates multi-scale contextual information and prepares the features for subsequent depth-aware fusion.

### *3.3.2. The Depth-Aware Attention Fusion (DAAF)*

This module takes RGB features as the primary information carrier and uses depth features as attention-guiding signals. Through a dual-branch cross-modal attention mechanism, it enhances structural awareness and ultimately performs adaptive fusion with the input features via learnable weighting factors. This ensures that the fused features retain the color and texture information from the RGB input while incorporating the structural and spatial constraints from the depth features. The overall architecture is illustrated in Fig. 5.

The RGB features extracted by MbAM module are represented as $F_R \in \mathbb{R}^{C\times H\times W}$. The depth information $D$obtained from the pretext task is processed through a simple convolution layer $\varepsilon$ to produce $D' \in \mathbb{R}^{C_d\times H\times W}$. $F_R$ provides rich texture and semantic information, while $D'$ primarily encodes the geometric structure and spatial relationships of the scene.

In DAAF, $F_R$ is first projected via a 1×1convolution and decomposed into the query, key, and value components $Q_R$, $K_R$and $V_R$. Considering that different attention branches emphasize distinct aspects of feature processing, and that different subspaces within the RGB features exhibit functional specialization for structure enhancement and detail recovery, $V_R$ is further decomposed into two complementary subspaces, $V_R^{\alpha}$ and $V_R^{\beta}$, via two parallel 1×1 convolutional projections.

The decomposition of $V_R$ realizes a decoupled bidirectional cross-modal modulation, allowing appearance-guided attention and geometry-guided attention to operate independently on complementary RGB subspaces. Specifically, $V_R^{\alpha}$ is designed as an RGB subspace guided by structural consistency, where the RGB features are modulated by depth-aware queries to strengthen appearance responses with spatial coherence under explicit structural constraints. On the other hand, $V_R^{\beta}$ serves as an appearance-aware RGB subspace, injecting photometric and texture cues into the depth-guided attention to alleviate over-smoothing issues that may arise from relying solely on structural constraints.

In contrast, the depth feature $D'$ is projected only into the $Q_d$ and $K_d$ components, without constructing a corresponding value component. This design is motivated by the fact that depth information primarily provides structural constraints and spatial guidance in the model, rather than supplying discriminative visual semantic information (e.g. object classes or texture patterns) relevant to the task objective. By using only the $Q_d$ and $K_d$ components, the depth feature emphasizes the geometric properties of pixels or regions in 3D space, without directly injecting feature values in the form of a value component. This avoids the explicit propagation of potential noise or errors from depth estimation into the RGB feature space. Consequently, this strategy enhances structural consistency while effectively preserving the semantic integrity and stability of the RGB features, rendering the cross-modal fusion process more robust.

In the RGB branch, the query component of the depth feature $Q_d$ is multiplied with the key component $K_R$ of the RGB feature to construct the cross-modal attention relationship:

$$A_{RGB} = \text{Softmax}(\frac{Q_d K_R^T}{\sqrt{C/8}}) \quad (15)$$

The RGB attention weight map $A_{RGB}$encodes the correlations of the RGB features from a depth-aware perspective, highlighting regions with structural consistency under geometric guidance. Subsequently, this attention is applied to the RGB subspace $V_R^{\alpha}$, producing the depth-enhanced RGB feature $f_R$: $f_R =$

$A_{RGB} \otimes V_x^{\alpha}$. This branch primarily enforces semantic consistency of RGB features under depth-guided structural constraints while capturing global dependencies.

In the depth branch, the query component of the RGB feature $Q_R$ and the key component of the depth feature $K_d$ are used to construct the attention map:

$$A_D = \text{Softmax}(\frac{Q_R K_d^T}{\sqrt{C/8}}) \tag{16}$$

The depth attention weight map $A_D$ captures appearance-guided geometric correlations, enabling RGB cues to modulate the response of depth-aware attention. It is then applied to the other RGB subspace $V_R^{\beta}$, producing the RGB feature $f_D$: $f_D = A_D \otimes V_x^{\beta}$,which integrates depth information while enhancing structural awareness and boundary constraints.

After fusing the outputs of the two branches, a 1×1 convolution followed by normalization (denoted as $\Phi$) is applied to integrate the features:

$$f_t = \Phi(f_R + f_D) \tag{17}$$

To prevent the attention-enhanced features from excessively perturbing the original representations and to improve training stability, DAAF introduces two learnable scaling parameters, $\hat{\eta}$ and $\hat{\mu}$, to adaptively balance the input features and the enhanced features:

$$f_o = \hat{\eta} \cdot f_t + \hat{\mu} \cdot F_R \tag{18}$$

### *3.4. Loss function*

Our framework is trained end-to-end, and the loss function consists of four parts:

$$\mathcal{L} = \lambda_1 \mathcal{L}_{L1}(I_{en}, I_n) + \lambda_2 \mathcal{L}_{per}(I_{en}, I_n) + \lambda_3 \mathcal{L}_{SSIM}(I_{en}, I_n) + \lambda_4 \mathcal{L}_{grad}(I_{en}, I_l) \tag{19}$$

Here $\mathcal{L}_{L1}$, $\mathcal{L}_{pre}$, $\mathcal{L}_{pre}$ and $\mathcal{L}_{pre}$represents the L1 reconstruction loss, perceptual loss [57], Structural similarity loss and gradient consistency loss respectively. The weighting factors are set as 1.0, 0.5, 0.1 and 0.3 respectively.

## 4. Experiment

In this section, we first describe the implementation details, evaluation datasets, and performance metrics. Next, we present quantitative and qualitative comparisons with state-of-the-art methods. Finally, ablation studies are conducted to validate the contribution of each component.

### *4.1. Experimental details*

#### *4.1.1. Implementation details*

We implemented our framework in PyTorch and conducted experiments on an NVIDIA A100 GPU. The model parameters are updated using the Adam optimizer ($\beta_1 = 0.9, \beta_2 = 0.999$), with an initial learning rate of $5 \times 10^{-5}$and a weight decay of $1 \times 10^{-5}$to enhance generalization. Training is performed for 200 epochs with a fixed batch size of 4, and input images are resized to 256×256 pixels. The learning rate is dynamically adjusted using a cosine annealing scheduler, with the minimum learning rate set to $5 \times 10^{-9}$. Gradient clipping with a threshold of 0.1 is applied to ensure training stability.

**Table 1**

Quantitative results of different methods on the LOL and SICEv2 datasets in terms of PSNR, SSIM [58], and LPIPS [59]. The 'Strategy' column indicates the training paradigm adopted by each method, where 'S' denotes supervised learning and 'U' denotes unsupervised learning. bold text indicates the **best result**, and underlined text indicates the second-best result.

| Datasets / Methods | Strategy | LOL-V1 | LOL-V2-real | LOL-V2-Synthetic | SICEv2 |
|---|---|---|---|---|---|
| | | PSNR ↑ / SSIM ↑ / LPIPS ↓ | PSNR ↑ / SSIM ↑ / LPIPS ↓ | PSNR ↑ / SSIM ↑ / LPIPS ↓ | PSNR ↑ / SSIM ↑ / LPIPS ↓ |
| RetinexNet[41] | S | 18.834 / 0.423 / 0.453 | 15.998 / 0.407 / 0.544 | 17.137 / 0.753 / 0.267 | 16.334 / 0.536 / 0.422 |
| EnlightenGAN[60] | U | 17.477 / 0.660 / 0.331 | 18.701 / 0.673 / 0.709 | 16.566 / 0.781 / 0.211 | 17.581 / 0.718 / 0.410 |
| KinD[61] | S | 17.641 / 0.773 / 0.169 | 19.334 / 0.827 / 0.343 | 17.265 / 0.793 / 0.256 | 17.988 / 0.806 / 0.253 |
| KinD++[39] | S | 17.793 / 0.769 / 0.182 | 17.735 / 0.774 / 0.207 | 17.457 / 0.806 / 0.241 | 18.019 / 0.812 / 0.241 |
| ZeroDCE[37] | U | 14.853 / 0.573 / 0.430 | 18.013 / 0.577 / 0.314 | 17.743 / 0.818 / 0.168 | 16.870 / 0.669 / 0.304 |
| ZeroDCE++[38] | U | 15.285 / 0.590 / 0.353 | 18.266 / 0.554 / 0.311 | 15.573 / 0.815 / 0.183 | 16.375 / 0.653 / 0.282 |
| SCI[47] | U | 13.812 / 0.526 / 0.346 | 17.258 / 0.546 / 0.368 | 16.682 / 0.746 / 0.257 | 15.859 / 0.606 / 0.236 |
| LLFlow[35] | S | 20.034 / 0.844 / 0.146 | 17.635 / 0.854 / **0.131** | 21.713 / 0.893 / 0.155 | 19.725 / 0.864 / 0.134 |
| PairLIE[45] | U | 18.979 / 0.748 / 0.274 | 19.824 / 0.785 / 0.273 | 19.024 / 0.704 / 0.230 | 19.276 / 0.746 / 0.258 |
| RUAS[34] | U | 10.664 / 0.314 / 0.335 | 12.736 / 0.475 / 0.348 | 13.536 / 0.650 / 0.319 | 12.417 / 0.480 / 0.336 |
| Retinexformer[14] | S | 23.155 / 0.813 / 0.183 | 21.171 / 0.896 / 0.181 | 10.119 / 0.710 / 0.274 | 18.148 / 0.806 / 0.213 |
| LLFormer[62] | S | 23.257 / 0.837 / 0.169 | 21.468 / 0.877 / 0.137 | 18.132 / 0.811 / 0.199 | 21.873 / 0.849 / 0.182 |
| EMNet[63] | S | 22.281 / 0.846 / 0.145 | 21.409 / 0.892 / 0.139 | 17.865 / 0.839 / 0.202 | 20.631 / 0.813 / 0.158 |
| SMG-LLIE[51] | **S** | **24.631** / 0.844 / 0.169 | 20.618 / 0.873 / 0.148 | **22.173** / 0.874 / 0.144 | 22.474 / 0.864 / 0.134 |
| SCRnet[64] | U | 23.174 / 0.838 / 0.203 | 21.756 / 0.844 / 0.152 | 20.982 / 0.861 / 0.151 | 21.946 / 0.852 / 0.167 |
| FIPNet[65] | S | 22.694 / 0.847 / 0.158 | 20.834 / 0.801 / 0.173 | 20.517 / 0.851 / 0.147 | 21.386 / 0.843 / 0.165 |
| Ours | S | 23.301 / **0.855** / **0.142** | **22.032** / **0.899** / 0.134 | 21.931 / **0.882** / **0.110** | **22.583** / **0.877** / **0.133** |

#### *4.1.2. Datasets and metrics*

We used low-light image pairs collected from the SICEv2 and LOL datasets to train our network. The LOL dataset has two versions: v1 [41] and v2 [48]. Compared to LOL-v1, which contains only real-world data, LOL-v2 is divided into a real-data subset and a synthetic-data subset. The SICEv2 dataset contains 512 low-light images.

Additionally, we selected 30 image pairs from SICEv2 and 15 image pairs from the official LOL evaluation set for performance evaluation. Since both datasets provide reference images, PSNR, SSIM [58], and LPIPS [59] were employed as evaluation metrics, where higher PSNR and SSIM values indicate closer similarity to the reference images, while lower LPIPS scores correspond to better enhancement quality. Furthermore, the superior performance of the proposed model has been extensively validated on multiple unpaired datasets, including MEF, LIME, and DICM. These unpaired datasets contain only low-light images without matching normal-light images.

### *4.2. Results analysis on paired datasets*

#### *4.2.1. Quantitative analysis*

Table 1 presents a systematic quantitative comparison between the proposed method and state-of-the-art approaches. To comprehensively validate the effectiveness of the proposed method, this study conducts comparative experiments with 15 representative algorithms, including 7 self-supervised methods and 9 fully supervised methods. The evaluation is performed from multiple perspectives to demonstrate the superiority and competitiveness of the proposed approach on the related tasks.

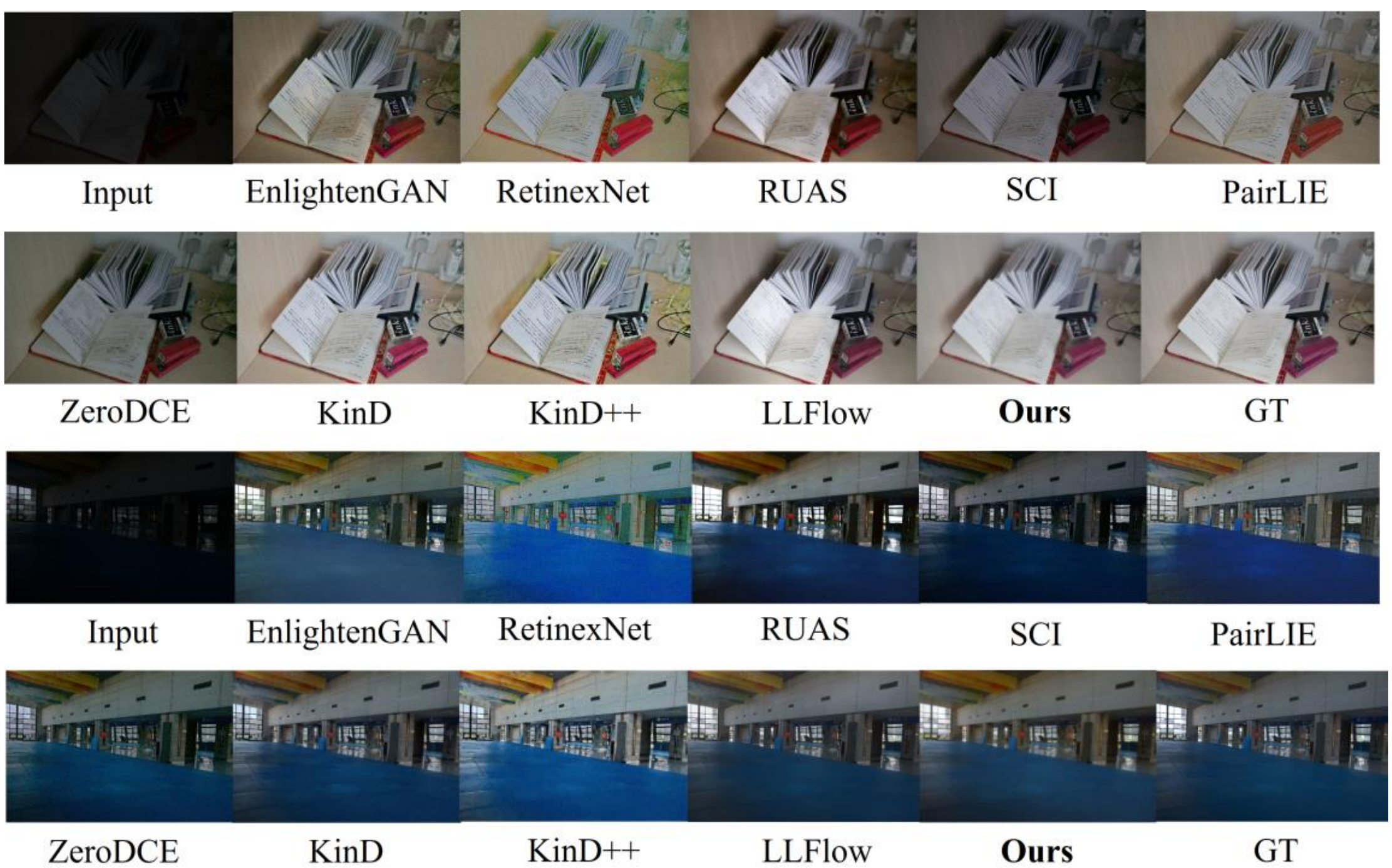

Fig. 6. On the LOLv1 dataset, we conduct comparative experiments between the proposed DMSA-Net and several representative advancing low-light image enhancement methods. Qualitative evaluation indicates that the proposed approach achieves superior visual consistency in terms of detail restoration, noise suppression, and color naturalness.

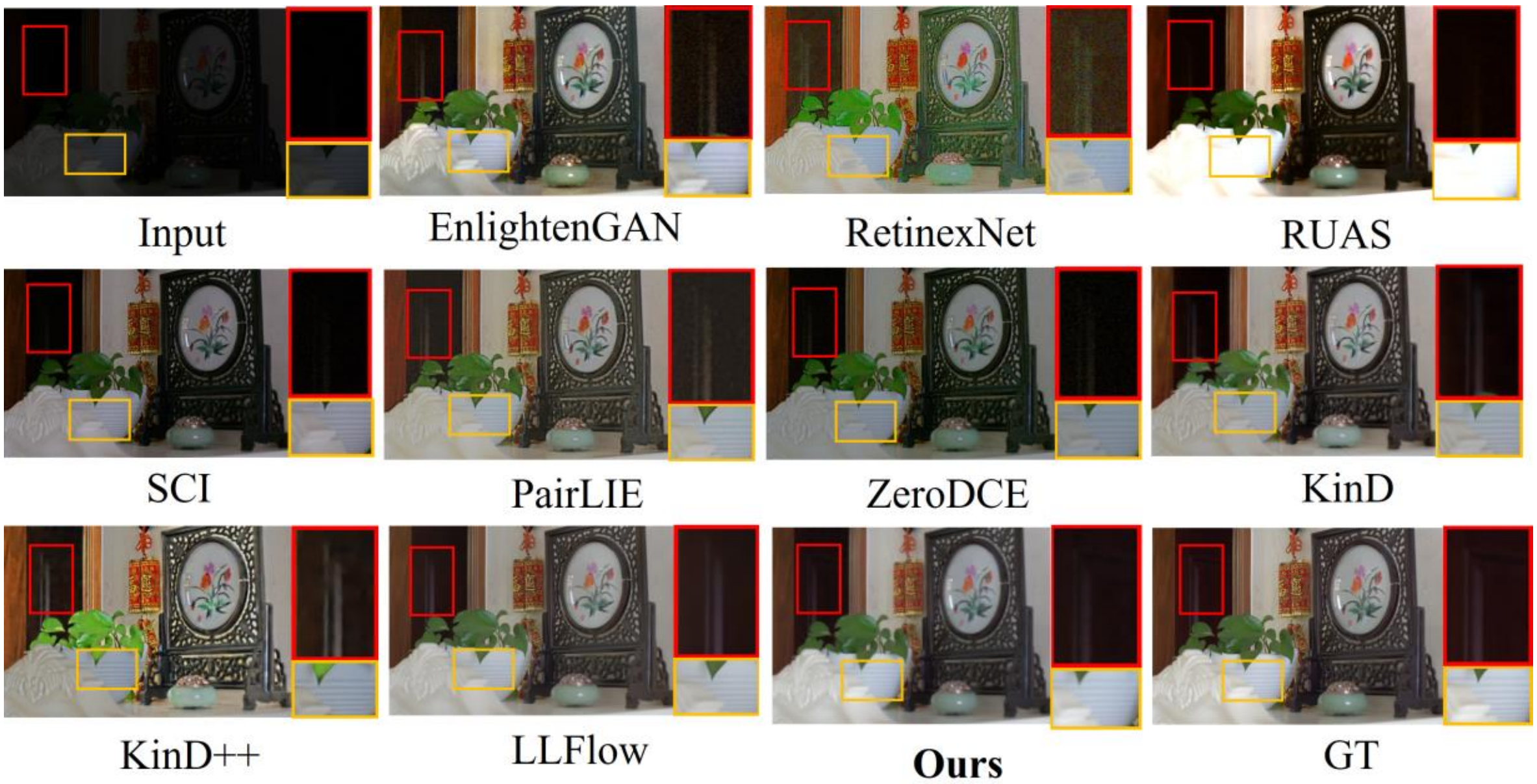

Fig. 7. Visual comparison of detail enhancement results between the proposed method (DMSA-Net) and other advancing low-light image enhancement methods on the LOLv1 dataset.

Experiments on the real low-light dataset LOL-V2-real demonstrate that our method achieves the best performance on the two key metrics, PSNR and SSIM, while obtaining an excellent LPIPS score of 0.137, which is close to the optimal method. This result verifies the stability and effectiveness of the proposed enhancement framework in complex real-world imaging conditions. On the synthetic dataset LOL-V2-Synthetic, our method shows slightly lower performance than the best competing approach in terms of SSIM. This phenomenon can be attributed to the characteristics of the synthetic dataset, where the low-light images are generated through algorithmic illumination degradation. Such synthetic distortions often favor methods that optimize strict pixel-level structural similarity. In contrast, the depth-guided mechanism in our framework mainly captures global spatial layout and scene geometry, which contributes more to perceptual coherence than to strict pixel-wise correspondence.

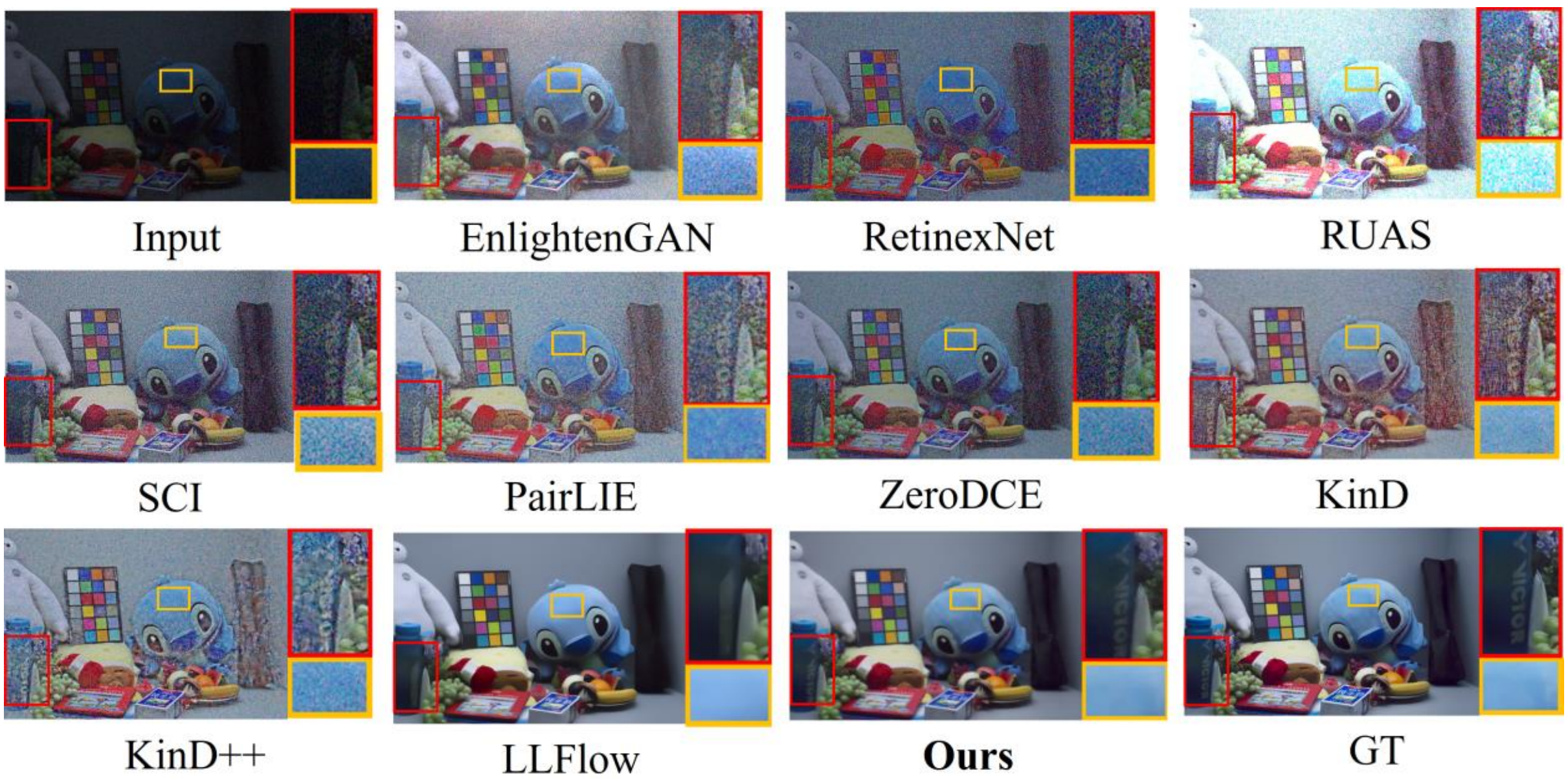

Fig. 8. Visual comparison of detail enhancement results between the proposed method (DMSA-Net) and other state-of-the-art low-light image enhancement methods on the LOLv2-real dataset.

Notably, however, on both the LOL-V2-Synthetic and the multi-exposure SICE datasets, our method achieves the best LPIPS performance, surpassing all compared methods in terms of perceptual quality. This result indicates that the spatial priors introduced by depth information effectively guide the reconstruction of semantically and perceptually consistent structures, leading to enhanced images that are more aligned with human visual perception rather than merely optimizing pixel-level similarity.

Considering the results across all four datasets, the proposed method achieves the best or second-best performance on most metrics, with particularly stable and outstanding performance on perceptual quality. These findings fully demonstrate the effectiveness of incorporating depth information as a geometric prior, as the provided spatial structure guidance significantly enhances the visual realism and naturalness of the enhanced images.

#### *4.2.2. Qualitative analysis*

To evaluate visual quality, we conducted qualitative comparisons between the proposed method and the baseline methods listed in Table 1 across multiple benchmark datasets, including LOLv1, LOL-v2-real, LOL-v2-Synthetic, and SICEv2. These results highlight differences in luminance restoration, detail preservation, and color naturalness.

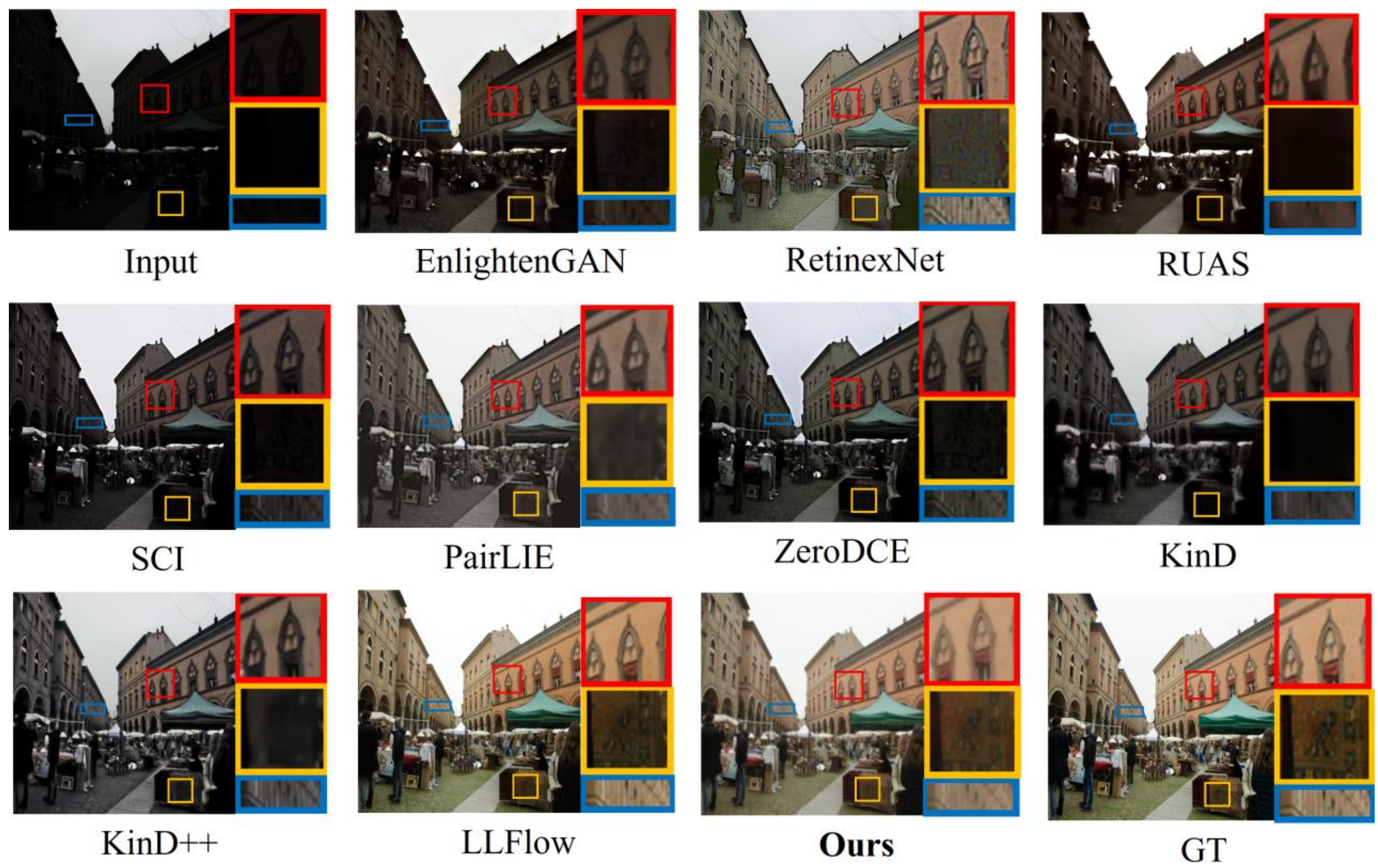

Fig. 9. Visual comparison of detail enhancement results between the proposed method (DMSA-Net) and other state-of-the-art low-light image enhancement methods on the LOLv2- Synthetic dataset.

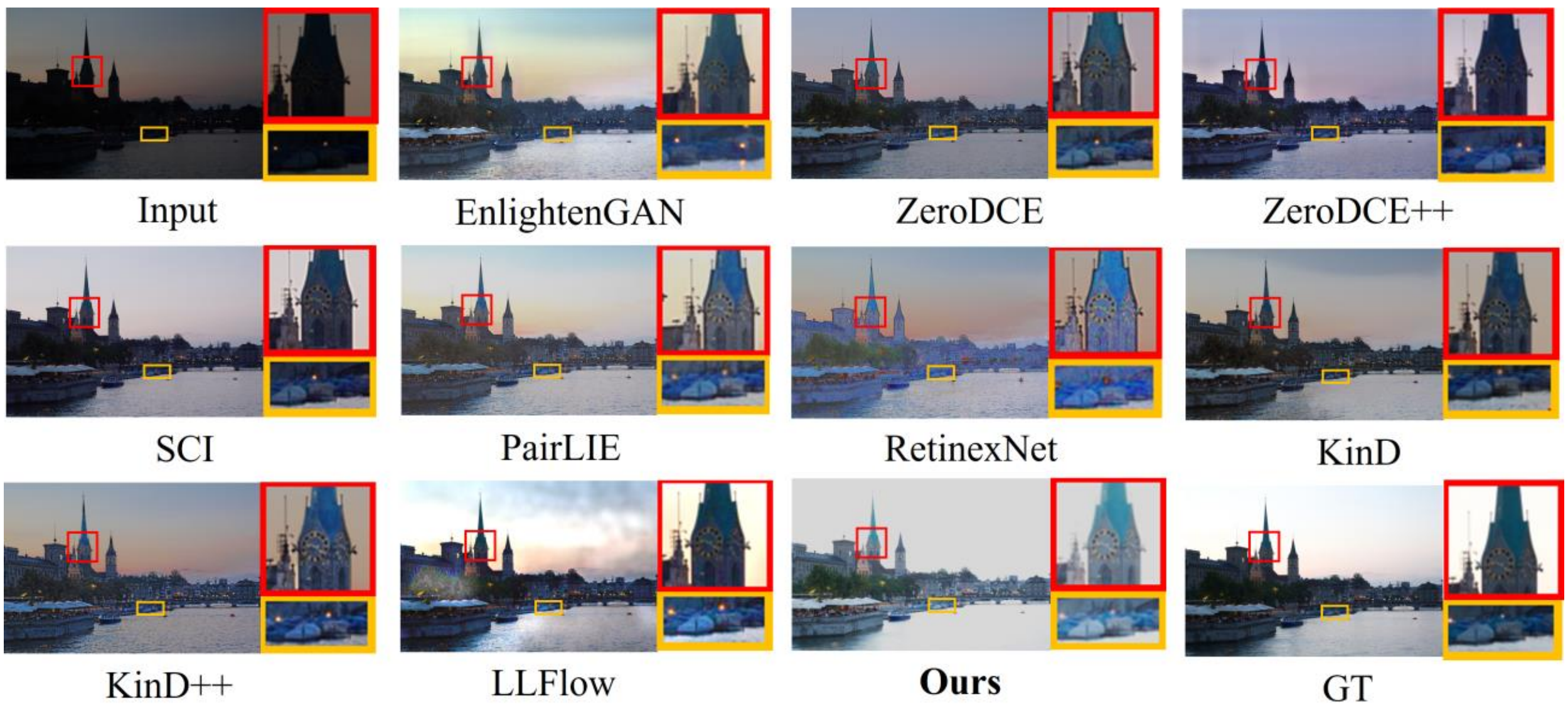

Fig. 10. Visual comparison of detail enhancement results between the proposed method (DMSA-Net) and other state-of-the-art low-light image enhancement methods on the SICEv2 dataset.

As illustrated in Fig. 6, we conduct a qualitative comparison between our DMSA-Net and several state-of-the-art methods. While earlier methods like RetinexNet [41] and EnlightenGAN [60] tend to introduce over-exposure or unnatural color shifts, and zero-shot approaches such as ZeroDCE [37] and SCI [47] often suffer from insufficient brightness enhancement, our DMSA-Net produces the most visually pleasing results. Specifically, the enhanced images from our method exhibit superior contrast and more realistic color reproduction, which are highly consistent with the ground truth (GT) and human visual preference. By effectively suppressing noise while restoring vivid textures, our method achieves a better balance between luminance enhancement and artifact control, demonstrating its robustness in handling diverse low-light scenarios.

Fig. 7 highlights the superior capability of our method in recovering structural details in regions with significant scene depth as well as in deeper dark areas. A key limitation of most existing supervised learning methods is their reliance on two-dimensional pixel-to-pixel mappings, which often leads to texture blurring or structural collapse in regions where depth variations are pronounced. As illustrated in the zoomed-in regions, many advanced supervised models struggle to recover clear object boundaries in distant areas with low signal-to-noise ratios. Our model explicitly incorporates spatial layout information through the Depth-Aware Attention Fusion (DAAF) and the Multi-scale Attention Module (MbAM). This depth-informed supervision enables our method to more effectively distinguish foreground from background objects, thereby ensuring high-fidelity recovery of fine details in deeper regions while maintaining minimal noise. The visualization results clearly validate the advantages of the proposed depth-aware strategy over conventional fully supervised paradigms that lack integrated spatial priors.

Fig. 8 illustrates a more challenging tabletop scene from the LOL-v2-real dataset, where most of the background remains under extremely low illumination. In such a high-noise environment, the majority of fully supervised baseline methods tend to produce severe color distortion and blotchy artifacts when attempting to recover details from shadowed regions, as highlighted in the zoomed-in red and yellow areas of the figure. Although LLFlow [35] is capable of generating smoother results, this improvement is typically achieved at the expense of fine textual details on object labels during the denoising process. In contrast, the proposed model achieves a favorable balance between effective noise suppression and the preservation of high-frequency details. By leveraging the Multi-scale Attention Module (MbAM), the network is able to capture both the local texture of the plush toy and the global illumination distribution across the scene. Compared with other supervised approaches, our model produces images with higher contrast and more accurate color reproduction, demonstrating its superior capability in addressing real-world degradation by effectively integrating spatial and multi-scale prior information.

As shown in Fig. 9, we further evaluate the performance of our model on the LOL-v2-Synthetic dataset. This dataset contains a variety of outdoor scenes characterized by illumination transitions and multiple depth layers. To rigorously assess the model's ability to preserve structural integrity across different spatial scales, we select a representative scene and analyze three distinct depth regions: foreground textures (yellow box), mid-range windows (red box), and distant structures (blue box). In the mid-range and distant regions, most existing methods struggle to balance contrast enhancement with the preservation of clear structural features. Some models prioritize fine texture representation but fail to properly regulate illumination, leading to noticeable color shifts in the restored scenes. Moreover, in the foreground texture region (yellow box), although several methods can recover basic visibility, they often exhibit color distortion or inconsistent local contrast. In contrast, our method is capable of capturing local details while maintaining global color harmony. The resulting enhanced images demonstrate superior perceptual quality, with the textures within the yellow box appearing noticeably finer and more coherent than those produced by competing methods.

As illustrated in Fig. 10, our model also demonstrates excellent global luminance balancing capability in high dynamic range (HDR) scenes from the SICEv2 dataset, while effectively preserving fine structural details. We select a representative outdoor scene—a cityscape featuring a distant clock tower—to evaluate the model's performance across different depth layers. In the distant building region (red box) and the mid-range region containing boats on the water surface (blue box), several existing methods introduce color distortions or halo artifacts. In contrast, our approach faithfully restores the natural colors surrounding the objects and preserves richer texture details of the boats. These qualitative results further confirm that our method outperforms competing approaches in scene-aware enhancement, demonstrating a clear advantage in maintaining spatial consistency and geometric fidelity in large real-world

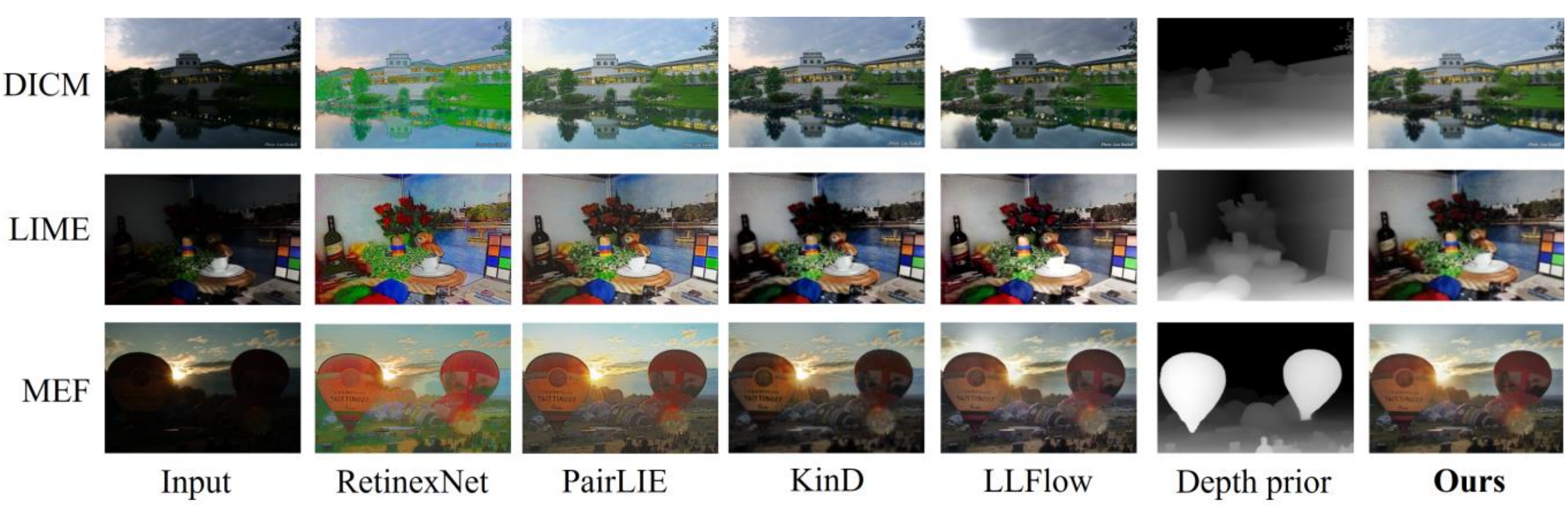

Fig. 11. Visual comparison of detail enhancement results between the proposed method (DMSA-Net) and other state-of-the-art low-light image enhancement methods on the unpair dataset.

**Table 2**

Quantitative results of different methods on unpaired datasets in terms of NIQE, BRISQUE, LOE, and UCIQE. Bold text indicates **the best result**, while underlined text denotes the second-best result.

| Methods \ Datasets | DICM | LIME | MEF |
|---|---|---|---|
| | NIQE ↓ / BRISQUE ↓ / LOE ↓ / UCIQE ↑ | NIQE ↓ / BRISQUE ↓ / LOE ↓ / UCIQE ↑ | NIQE ↓ / BRISQUE ↓ / LOE ↓ / UCIQE ↑ |
| Input | 4.995 / 34.690 / - / 25.814 | 10.304 / 19.799 / - / 27.663 | 10.629 / 34.052 / - / 18.580 |
| RetinexNet[41] | 5.622 / 32.487 / 75.069 / 14.747 | 7.558 / 17.714 / 83.588 / 29.178 | 9.754 / 31.157 / 74.402 / 20.301 |
| PairLIE[45] | 4.499 / 28.328 / 68.983 / 28.401 | 6.154 / 10.649 / 82.790 / 21.877 | 6.201 / 25.358 / 63.915 / **24.701** |
| KinD[61] | 4.502 / 27.929 / 45.008 / 26.116 | 4.947 / 12.320 / 69.182 / 34.467 | 5.431 / 25.984 / **43.420** / 23.016 |
| LLFlow[35] | **4.331** / 28.917 / 56.993 / 27.436 | 5.008 / **10.202** / 80.423 / **44.586** | 6.441 / 22.825 / 62.056 / 21.422 |
| Ours | 4.447 / **26.701** / **38.477** / **28.882** | **4.513** / 10.477 / **60.014** / 37.298 | **6.063** / **19.418** / 57.010 / 23.876 |

scenes. The depth-guided design ensures that distant structures retain clear contours and higher contrast, effectively avoiding the structural blurring commonly observed in other methods.

### *4.3. Results analysis on unpaired datasets*

To objectively evaluate the performance of the proposed algorithm for low-light image enhancement, we conduct comprehensive quantitative comparison experiments on three unpaired datasets, namely DICM, LIME, and MEF, using a model trained on the LOLv2 dataset. Table 2 presents the performance comparison between the proposed method (Ours) and five representative state-of-the-art algorithms across four key evaluation metrics, including NIQE [66], BRISQUE [67], LOE [68], and UCIQE [69]. Fig. 11 provides the visual comparison results of different methods, as well as the depth priors extracted by our pretext task module. The experimental results demonstrate that the proposed method exhibits significant competitive advantages across multiple evaluation dimensions.

Regarding the UCIQE metric, which reflects the balance between image contrast and color saturation, our method achieves the best score of 28.882 on the DICM dataset, and obtains the second-best scores of 37.298 and 23.876 on the LIME and MEF datasets, respectively. This indicates that the color restoration mechanism adopted in our framework effectively suppresses the color cast commonly observed under low-light conditions, resulting in enhanced images that appear more vivid and perceptually pleasing. Meanwhile, the LOE analysis shows that our method achieves the lowest values on both the DICM and LIME datasets. This result strongly demonstrates the effectiveness of our optimization strategy in balancing global brightness mapping and local detail preservation, thereby avoiding issues such as local overexposure or brightness inversion during enhancement.

In terms of NIQE and BRISQUE, although the original images may exhibit relatively high statistical naturalness under certain distributions, their extremely low visibility significantly limits practical usability. Compared with other enhancement methods, our approach achieves the best NIQE score of 4.513 on the LIME dataset and substantially outperforms competing methods with a BRISQUE score of 19.418 on the MEF dataset. These results indicate that our algorithm is capable of increasing brightness in dark regions while introducing fewer artificial artifacts and preserving richer texture details.

As can also be observed from the visual comparison in Fig. 11, compared with existing methods, the proposed approach produces more visually pleasing results with better structural consistency. The results generated by RetinexNet [41] often suffer from color distortion and unnatural saturation, especially in the LIME example where objects appear overly vivid. PairLIE [45] improves brightness, but still exhibits insufficient contrast and a slight loss of details in dark regions. KinD [61] enhances illumination more effectively; however, some regions appear over-smoothed and lack fine texture details. LLFlow [35] achieves relatively balanced brightness but introduces slight haze and fails to fully restore local contrast in certain areas. In contrast, the proposed method restores natural brightness while preserving rich textures and clear structural boundaries across all scenes. In the DICM example, reflections and architectural details are more clearly recovered; in the LIME scene, object colors appear more natural without oversaturation; and in the MEF example, the sky gradient and balloon contours remain clear and visually coherent.

In summary, the consistent superiority in quantitative metrics, together with the corresponding improvements in qualitative visual results, jointly demonstrate the effectiveness and advanced capability of the proposed method in addressing the challenges of unpaired low-light image enhancement.

### *4.4. Comprehensive ablation and architectural validation*

To validate the effectiveness of each component in the proposed depth-aware multi-scale attention network, we conduct comprehensive ablation studies from two perspectives: progressive module ablation and objective architectural analysis. The former evaluates the contribution of each core component through progressive module insertion, while the latter further investigates several key architectural design choices, including depth guidance, fusion strategies, multi-scale modulation design, and the role of DAAF. All experiments are conducted on the LOL-v2-real dataset, with PSNR, SSIM, and LPIPS adopted as evaluation metrics.

#### *4.4.1. Progressive module ablation*

To validate the effectiveness of each component in the proposed depth-aware multi-scale attention network, we conducted systematic ablation studies. Key modules were progressively removed or replaced to analyze their individual contributions to the enhanced performance. All experiments were conducted on the LOL-V2-real dataset, with PSNR, SSIM, and LPIPS used as evaluation metrics. The configurations of the components in the ablation studies are as follows:

(a) Baseline model: Only the U-Net architecture is used, with low-light RGB images as input, without incorporating depth information;

(b) Baseline + direct depth concatenation: On the U-Net architecture, the depth map is directly concatenated with the RGB image along the channel dimension;

(c) Baseline + DAAF module: Based on the U-Net architecture, only the proposed DAAF module is applied;

(d) Baseline+ MbAM module: Based on the U-Net architecture, only the proposed MbAM is applied;

(e) Baseline + MbAM + DAAF modules: Based on the U-Net architecture, both the proposed MbAM and DAAF modules are incorporated.

(f) DMSA-Net: the complete model is used.

The final ablation results are shown in Table 3.

**Table 3**

Test results of different ablation experimental settings. Bold text indicates **the best result** for this indicator.

| Dataset / Methods | LOL-V2-Real | | |
|---|---|---|---|
| | PSNR ↑ | SSIM ↑ | LPIPS ↓ |
| a | 13.354 | 0.533 | 0.463 |
| b | 14.173 | 0.576 | 0.385 |
| c | 17.583 | 0.788 | 0.246 |
| d | 16.124 | 0.763 | 0.255 |
| e | 21.458 | 0.872 | 0.139 |
| **f** | **22.032** | **0.899** | **0.134** |

The experimental results demonstrate that the complete model achieves the best performance across all evaluation metrics, validating the effectiveness of each component. The baseline model provides only basic low-light restoration capabilities. Introducing depth information, whether via simple concatenation or the proposed DAAF, improves performance, with DAAF significantly outperforming direct concatenation due to its effective integration of depth and appearance features. The MbAM alone also substantially enhances performance, highlighting the importance of multi-scale contextual information in low-light enhancement. When DAAF and MbAM are combined, the model achieves further synergistic gains, indicating the complementary relationship between depth guidance and multi-scale feature extraction. Incorporating the RDB improves structural consistency and pixel-level reconstruction quality while having a limited impact on perceptual similarity, suggesting that RDB primarily refines local texture and structural details, whereas DAAF plays the core role in geometric guidance.

To further investigate the role of depth priors and the optimal strategy for integrating depth information, we conduct additional ablations on depth guidance and fusion location.

#### *4.4.2. Depth guidance and fusion strategy analysis*

(1) Depth or No-depth

First, we compare the baseline model without depth guidance and the full model with depth priors. As shown in Table 4, introducing depth information significantly improves all evaluation metrics, demonstrating the necessity of geometry-aware guidance.

(2) Different fusion strategies

We further compare three depth fusion strategies, including encoder-only fusion, decoder-only fusion, and encoder-decoder fusion. The results show that encoder-only fusion achieves the best performance. This is because encoder features are more compatible with depth priors, whereas decoder features are reconstruction-oriented, and introducing depth at this stage may lead to modality inconsistency and noise propagation.

#### *4.4.3. Multi-scale modulation and DAAF analysis*

To validate the effectiveness of the proposed feature interaction modules, we further analyze the impact of MFMA and DAAF.

(1) MFMA or simple blocks

**Table 4**

Ablation analysis of depth guidance and depth fusion strategies on LOL-V2-real. Bold text indicates **the best result** for this indicator.

| Configuration \ Metrics | Depth Guidance | Fusion Strategy | PSNR ↑ | SSIM ↑ | LIPIS ↓ |
|---|---|---|---|---|---|
| *Baseline* | × | - | 16.201 | 0.647 | 0.393 |
| *Ours w/ depth* | √ | Encode | **21.976** | **0.895** | **0.131** |
| *Decoder-only fusion* | √ | Decode | 20.758 | 0.833 | 0.174 |
| *Encoder and Decoder fusion* | √ | Encode-Decode | 21.019 | 0.871 | 0.156 |

**Table 5**

Ablation analysis of MFMA and DAAF modules. Bold text indicates **the best result** for this indicator.

| Configuration \ Metrics | MFMA | DAAF | PSNR ↑ | SSIM ↑ | LIPIS ↓ |
|---|---|---|---|---|---|
| *Simple multi-scale concat* | × | √ | 19.836 | 0.812 | 0.279 |
| *Standard convolution fusion* | × | √ | 20.034 | 0.831 | 0.253 |
| *w/o DAAF* | √ | × | 21.015 | 0.828 | 0.196 |
| *Full model (Ours)* | √ | √ | **22.032** | **0.899** | **0.134** |

**Table 6**

Quantitative analysis results of the plug-and-play DAAF module on different low-light image enhancement algorithms on the LOLv2-real dataset. Bold text indicates **the best result** for this indicator.

| Methods \ Datasets | LOL-V2-real | SICEv2 |
|---|---|---|
| | PSNR ↑ / SSIM ↑ / LPIPS ↓ | PSNR ↑ / SSIM ↑ / LPIPS ↓ |
| KinD[60] | 19.334 / 0.827 / 0.343 | 17.988 / 0.806 / 0.253 |
| KinD+DAAF | **19.852** / **0.831** / **0.296** | **18.112** / **0.835** / **0.201** |
| RetinexNet[41] | 15.998 / 0.407 / 0.544 | 16.334 / 0.536 / 0.422 |
| RetinexNet+DAAF | **17.013** / **0.612** / **0.373** | **17.293** / **0.641** / **0.333** |
| LLFLow[35] | **17.635** / **0.854** / 0.131 | 19.725 / 0.864 / 0.134 |
| LLFLow+DAAF | 17.508 / 0.846 / **0.129** | **20.064** / **0.883** / **0.120** |

As reported in Table 5, replacing MFMA with simpler multi-scale aggregation strategies (i.e., concat & convolution fusion) consistently degrades performance, confirming that explicit cross-scale modulation is more effective than conventional feature fusion.

(2) With or without DAAF

In addition, removing DAAF leads to noticeable drops across all metrics, especially in perceptual quality. This demonstrates that adaptive geometry-guided attention plays a critical role in effective RGB-depth fusion.

Overall, these results confirm that both MFMA and DAAF are indispensable components of the proposed architecture.

### *4.5. Plug-and-Play DAAF Module Experiment*

We select several representative low-light image enhancement algorithms as baseline models. Without modifying their original network architectures or training strategies, the proposed DAAF module and depth information are directly integrated into the corresponding models for training. Comparative experiments are then conducted to validate the generality and effectiveness of the proposed plug-and-play DAAF module. The corresponding qualitative visualization results are shown in Figs. 12 and 13, while the quantitative evaluation results are summarized in Table 6.

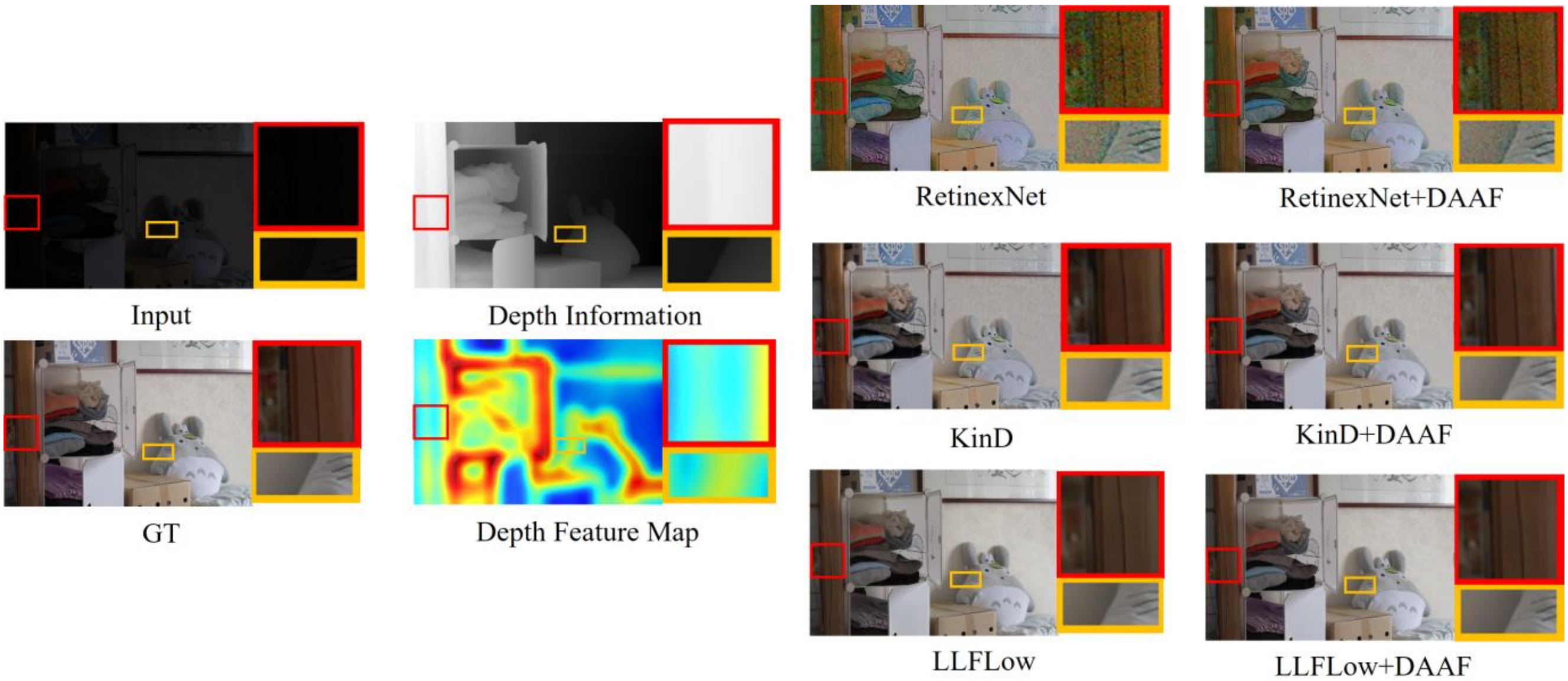

Fig. 12. Visual experiments of the plug-and-play DAAF module on different low-light image enhancement algorithms.

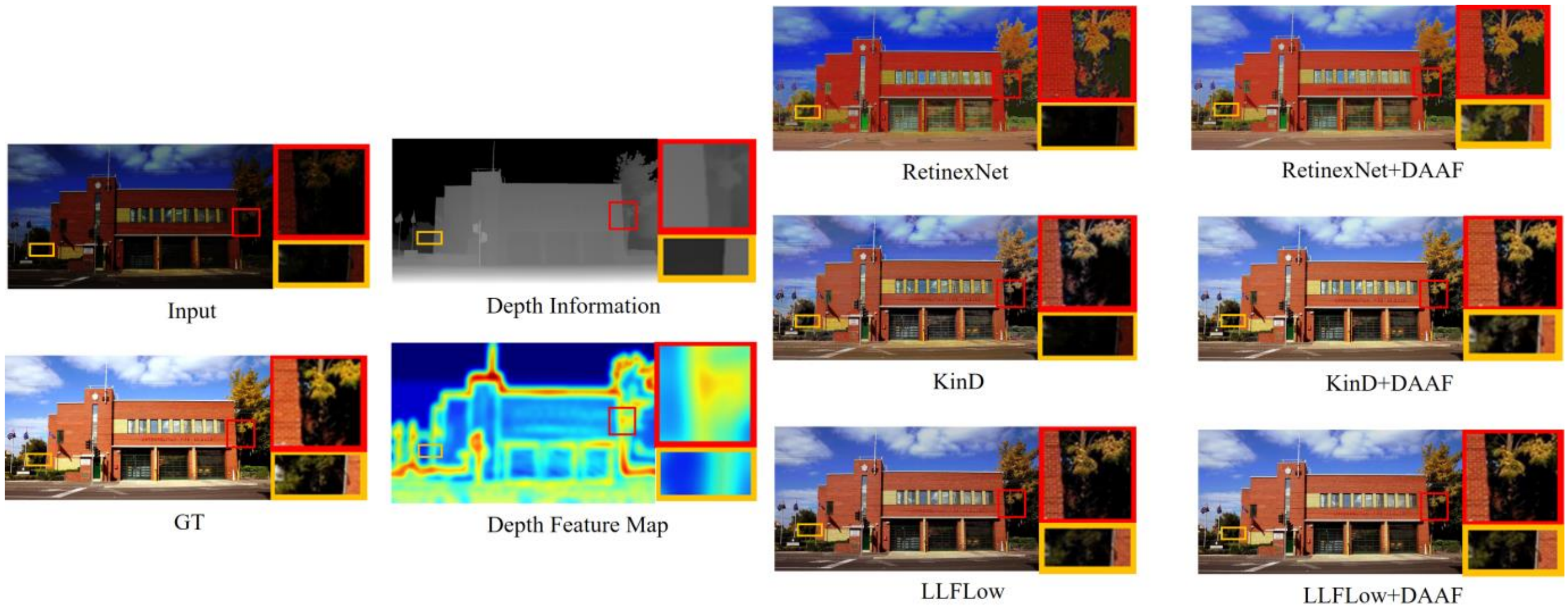

Fig. 13. Visual experiments of the plug-and-play DAAF module on different low-light image enhancement algorithms.

From a subjective visual perspective, the integration of the DAAF module consistently improves the performance of the baseline methods in terms of brightness balance, detail restoration, and color naturalness, regardless of whether the scene involves indoor low-light conditions or complex outdoor illumination. Dark regions are effectively enhanced, while bright regions do not exhibit noticeable overexposure or color distortion. Texture and structural details become clearer, and the overall visual appearance is closer to natural real-world illumination.

The quantitative results further support these observations: across multiple datasets, the introduction of the DAAF module consistently improves PSNR and SSIM while reducing the LPIPS score. This indicates that the DAAF module not only significantly enhances perceptual quality but also improves pixel-level accuracy and structural consistency. Specifically, baseline methods such as RetinexNet [41] and KinD [60] benefit from more substantial performance gains. Even for high-performing methods such as LLFlow [35], the DAAF module still provides moderate overall improvements, demonstrating its strong generalization capability and practical value as a plug-and-play component.

## 5. Conclusion

This paper presented **DMSA-Net**, a depth-guided multi-scale attention framework for geometry-aware low-light image enhancement. The proposed method addresses the limitations of appearance-driven enhancement models by introducing depth-related structural priors into low-light representation learning. A Retinex-based decomposition strategy is first used to obtain illumination-invariant reflectance representations, from which more reliable scene geometry can be inferred under degraded illumination. Then, a multi-scale depth-guided fusion mechanism is integrated into a hierarchical encoder–decoder architecture to adaptively combine geometric and appearance features. Through depth-aware feature recalibration, DMSA-Net improves both visual restoration quality and structural consistency. Experiments on multiple benchmark datasets demonstrate the effectiveness of the proposed framework in low-light restoration and structure preservation. Furthermore, the constructed **LOL-D** dataset provides a depth-augmented benchmark for geometry-aware low-light vision research. This work offers a useful exploration of depth-guided representation learning for robust visual understanding under challenging illumination conditions.

## References

[1] M. Afifi, K. G. Derpanis, B. Ommer, and M. S. Brown, “Learning multi-scale photo exposure correction,” in Proc. IEEE/CVF Conf. Comput. Vis. Pattern Recognit. (CVPR), Nashville, TN, USA, 2021, pp. 9157–9167.

[2] M. Salah, et al., “NeuTac: Zero-shot sim2real measurement for neuromorphic vision-based tactile sensors,” IEEE Trans. Instrum. Meas., vol. 73, Sep. 2024, Art. no. 5031315.

[3] M. A. Hasan et al., “Image classification using convolution-al neural networks,” Int. J. Mech. Eng. Res. Technol., vol. 16, no. 2, pp. 155–162, Apr. 2024.

[4] D. Peng, W. Ding, and T. Zhen, “A novel low light object detection method based on the YOLOv5 fusion feature enhancement,” Sci. Rep., vol. 14, no. 1, Feb. 2024, Art. no. 4486.

[5] Y.-H. Wu et al., “Low-resolution self-attention for semantic segmentation,” IEEE Trans. Pattern Anal. Mach. Intell., vol. 47, no. 9, pp. 8180-8192, Jun. 2025.

[6] M. Abdullah-Al-Wadud, M. H. Kabir, M. A. A. Dewan, and O. Chae, “A dynamic histogram equalization for image contrast enhancement,” IEEE Trans. Consum. Electron., vol. 53, no. 2, pp. 593–600, May. 2007.

[7] H. Ibrahim and N. S. P. Kong, “Brightness preserving dynamic histogram equalization for image contrast enhancement,” IEEE Trans. Consum. Electron., vol. 53, no. 4, pp. 1752–1758, Nov. 2007.

[8] E. H. Land, “The retinex theory of color vision,” Sci. Am., vol. 237, no. 6, pp. 108–129, Dec. 1977.

[9] X. Fu, Y. Liao, D. Zeng, Y. Huang, X. Zhang, and X. Ding, “A probabilistic method for image enhancement with simultaneous illumination and reflectance estimation,” IEEE Trans. Image Process., vol. 24, no. 12, pp. 4965–4977, Aug. 2015.

[10] X. Guo, Y. Li, and H. Ling, “Lime: Low-light image enhancement via illumination map estimation,” IEEE Trans. Image Process., vol. 26, no. 2, pp. 982–993, Feb. 2017.

[11] M. Li, J. Liu, W. Yang, X. Sun, and Z. Guo, “Structure-revealing low-light image enhancement via robust retinex model,” IEEE Trans. Image Process., vol. 27, no. 6, pp. 2828–2841, Feb. 2018.

[12] S. Park, S. Yu, B. Moon, S. Ko, and J. Paik, “Low-light image enhancement using variational optimization-based retinex model,” IEEE Trans. Consum. Electron., vol. 63, no. 2, pp. 178–184, May. 2017.

[13] S. Wang, J. Zheng, H.-M. Hu, and B. Li, “Naturalness preserved enhancement algorithm for non-uniform illumination images,” IEEE Trans. Image Process., vol. 22, no. 9, pp. 3538–3548, Sep. 2013.

[14] Y. Cai, H. Bian, J. Lin, H. Wang, R. Timofte, and Y. Zhang, “Retinexformer: One-stage retinex-based transformer for low-light image enhancement,” in Proc. IEEE/CVF Int. Conf. Comput. Vis. (ICCV), Paris, France, 2023, pp. 12504–12513.

[15] Y.-S. Chen, Y.-C. Wang, M.-H. Kao, and Y.-Y. Chuang, “Deep photo enhancer: Unpaired learning for image enhancement from photographs with gans,” in Proc. IEEE Conf. Comput. Vis. Pattern Recognit. (CVPR), Salt Lake City, UT, USA, 2018, pp. 6306–6314.

[16] M. Z. U. Abideen et al., “SNRD-Net: SNR-aware dual enhancement network for low-light images,” Comput. Vis. Image Underst., vol. 268, Art. no. 104776, 2026.

[17] J. Tan, S. Pei, L. Huang, and K. Cong, “An efficient three-stage network via Multi-Scale Orthogonal Complementary Transformer for low-light image enhancement,” Comput. Vis. Image Underst., vol. 268, Art. no. 104778, 2026.

[18] H. Cui, et al., “Progressive dual-branch network for low-light image enhancement,” IEEE Trans. Instrum. Meas., vol. 71, 2022, Art. no. 2520318.

[19] H. Sun, D. Gao, P. He, X. Li, and F. Wang, “LLIE-Face: A multi-modal dataset for low-light facial image enhancement,” Comput. Vis. Image Underst., vol. 263, Art. no. 104576, 2026.

[20] X. Xu, R. Wang, C.-W. Fu, and J. Jia, “SNR-aware low-light image enhancement,” in Proc. IEEE/CVF Conf. Comput. Vis. Pattern Recognit. (CVPR), New Orleans, LA, USA, 2022, pp. 17714–17724.

[21] Y. Wu, F. Liu, and R. Wang, “JSF: A joint spatial-frequency domain network for low-light image enhancement,” Comput. Vis. Image Underst., vol. 261, Art. no. 104496, 2025.

[22] W. Dong et al., “Towards Scale-Aware Low-Light Enhancement via Structure-Guided Transformer Design,” in Proc. Comput. Vis. Pattern Recognit. Conf. (CVPR), Nashville City, TN, USA, 2025, pp. 1460-1470.

[23] H. Zhou et al., “Low-light image enhancement via generative perceptual priors,” in Proc. AAAI Conf. Artif. Intell. (AAAI), vol. 39, no. 10, 2025, pp. 10752-10760.

[24] Y. Lin et al., “Geometric-aware low-light image and video enhancement via depth guidance,” IEEE Trans. Image Process., vol. 34, pp. 5442-5457, Aug. 2025.

[25] C. Lee, C. Lee, and C.-S. Kim, “Contrast enhancement based on layered difference representation of 2d histograms,” IEEE Trans. Image Process., vol. 22, no. 12, pp. 5372–5384, Sep. 2013.

[26] C. Li, J. Guo, F. Porikli, and Y. Pang, “Lightennet: a convolutional neural network for weakly illuminated image enhancement,” Pattern Recognit. Lett., vol. 104, pp. 15–22, Mar. 2018.

[27] T. Celik and T. Tjahjadi, “Contextual and variational contrast enhancement,” IEEE Trans. Image Process., vol. 20, no. 12, pp. 3431–3441, Dec. 2011.

[28] T. Arici, S. Dikbas, and Y. Altunbasak, "A histogram modification framework and its application for image contrast enhancement," IEEE Trans. Image Process., vol. 18, no. 9, pp. 1921–1935, Sep. 2009.

[29] S.-C. Huang, F.-C. Cheng, and Y.-S. Chiu, "Efficient contrast enhancement using adaptive gamma correction with weighting distribution," IEEE Trans. Image Process., vol. 22, no. 3, pp. 1032–1041, Oct. 2012.

[30] S. Rahman, M. M. Rahman, M. A.-A. Wadud, G. D. Al-Quaderi, and M. Shoyaib, "An adaptive gamma correction for image enhancement," EURASIP J. Image Video Process., 2016, article number 35, Oct. 2016.

[31] G. Li, B. Chen, C. Zhao, et al., "Osmamba: Omnidirectional spectral mamba with dual-domain prior generator for exposure correction," in Proc. Comput. Vis. Pattern Recognit. Conf. (CVPR), Nashville City, TN, USA, 2025, pp. 7480–7490.

[32] X. Fu, D. Zeng, Y. Huang, X. Zhang, and X. Ding, "A weighted variational model for simultaneous reflectance and illumination estimation," in Proc. IEEE/CVF Conf. Comput. Vis. Pattern Recognit. (CVPR), Las Vegas, NV, USA, 2016, pp. 2782–2790.

[33] J. Xu, Y. Hou, D. Ren, L. Liu, F. Zhu, M. Yu, H. Wang, and L. Shao, "Star: A structure and texture aware retinex model," IEEE Trans. Image Process., vol. 29, pp. 5022–5037, Mar. 2020.

[34] R. Liu, L. Ma, J. Zhang, et al., "Retinex-inspired unrolling with cooperative prior architecture search for low-light image enhancement," in Proc. IEEE/CVF Conf. Comput. Vis. Pattern Recognit. (CVPR), Nashville, TN, USA, 2021, pp. 10561–10570.

[35] Y. Wang, R. Wan, W. Yang, et al., "Low-light image enhancement with normalizing flow," in Proc. AAAI Conf. Artif. Intell. (AAAI), 2022, vol. 36, no. 3, pp. 2604–2612.

[36] Z. Li, F. Zhang, M. Cao, et al., "Real-time exposure correction via collaborative transformations and adaptive sampling," in Proc. IEEE/CVF Conf. Comput. Vis. Pattern Recognit. (CVPR), Seattle City, WA, USA, 2024, pp. 2984–2994.

[37] C. Guo, C. Li, J. Guo, et al., "Zero-reference deep curve estimation for low-light image enhancement," in Proc. IEEE/CVF Conf. Comput. Vis. Pattern Recognit. (CVPR), Seattle City, WA, USA, 2020, pp. 1780–1789.

[38] C. Li, C. Guo, and C. C. Loy, "Learning to enhance low-light image via zero-reference deep curve estimation," IEEE Trans. Pattern Anal. Mach. Intell., vol. 44, no. 8, pp. 4225–4238, Mar. 2021.

[39] Y. Zhang, X. Guo, J. Ma, et al., "Beyond brightening low-light images," Int. J. Comput. Vis., vol. 129, no. 4, pp. 1013–1037, Jan. 2021.

[40] K. G. Lore, A. Akintayo, and S. Sarkar, "LLNet: A deep autoencoder approach to natural low-light image enhancement," Pattern Recognit., vol. 61, pp. 650–662, Jan. 2017.

[41] C. Wei et al., "Deep retinex decomposition for low-light enhancement," arXiv preprint arXiv:1808.04560, 2018.

[42] F. Lv et al., "MBLLEN: Low-Light Image/Video Enhancement Using CNNs," in Br. Mach. Vis. Conf., London, UK, 2018, p. 4.

[43] Y. Feng et al., "Difflight: integrating content and detail for low-light image enhancement," in Proc. IEEE/CVF Conf. Comput. Vis. Pattern Recognit. (CVPR), Seattle City, WA, USA, 2024, pp. 6143-6152.

[44] Z. Zhang et al., "Deep color consistent network for low-light image enhancement," in Proc. IEEE/CVF Conf. Comput. Vis. Pattern Recognit. (CVPR), New Orleans City, LA, USA, 2022, pp. 1899-1908.

[45] Z. Fu et al., "Learning a simple low-light image enhancer from paired low-light instances," in Proc. IEEE/CVF Conf. Comput. Vis. Pattern Recognit. (CVPR), Vancouver City, BC, CA, 2023, pp. 22252-22261.

[46] K. Wu et al., “Cycle-retinex: Unpaired low-light image enhancement via retinex-inline cyclegan,” IEEE Trans. Multimed., vol. 26, pp. 1213–1228, May. 2023.

[47] L. Ma et al., “Toward fast, flexible, and robust low-light image enhancement,” in Proc. IEEE/CVF Conf. Comput. Vis. Pattern Recognit. (CVPR), New Orleans City, LA, USA, 2022, pp. 5637-5646.

[48] W. Yang et al., “Sparse gradient regularized deep retinex network for robust low-light image enhancement,” IEEE Trans. Image Process., vol. 30, pp. 2072–2086, Jan. 2021.

[49] S. Jin et al., “Darkvisionnet: Low-light imaging via RGB-NIR fusion with deep inconsistency prior,” in Proc. AAAI Conf. Artif. Intell. (AAAI), vol. 36, no. 1, 2022, pp. 1104-1112.

[50] S. Yang et al., “Implicit neural representation for cooperative low-light image enhancement,” in Proc. IEEE/CVF Int. Conf. Comput. Vis. (ICCV), Paris, FR, 2023, pp. 12918-12927.

[51] X. Xu, R. Wang, and J. Lu, “Low-light image enhancement via structure modeling and guidance,” in Proc. IEEE/CVF Conf. Comput. Vis. Pattern Recognit. (CVPR), Vancouver City, BC, CA, 2023, pp. 9893-9903.

[52] L. Yang et al., “Depth anything v2,” in Adv. Neural Inf. Process. Syst., vol. 37, 2024, pp. 21875–21911.

[53] L. Piccinelli et al., “UniDepth: Universal monocular metric depth estimation,” in Proc. IEEE/CVF Conf. Comput. Vis. Pattern Recognit. (CVPR), Seattle City, WA, USA, 2024, pp. 10106-10116.

[54] A. Bochkovskii et al., “Depth pro: Sharp monocular metric depth in less than a second,” arXiv preprint arXiv:2410.02073, 2024.

[55] J. Johnson, A. Alahi, and L. Fei-Fei, “Perceptual losses for real-time style transfer and super-resolution,” in Eur. Conf. Comput. Vis. (ECCV), Cham, Switzerland: Springer Int. Publ., 2016, pp. 694–711.

[56] A. Gupta et al., “Memory-efficient Transformers via Top-k Attention,” arXiv preprint arXiv:2106.06899, 2021.

[57] T. S. Tammina, “Transfer learning using VGG-16 with deep convolutional neural network for classifying images,” Int. J. Sci. Res. Publ. (IJSRP), vol. 9, no. 10, pp. 143–150, Oct. 2019.

[58] Z. Wang, A. C. Bovik, H. R. Sheikh, and E. P. Simoncelli, “Image quality assessment: from error visibility to structural similarity,” IEEE Trans. Image Process., vol. 13, no. 4, pp. 600–612, Apr. 2004.

[59] R. Zhang, P. Isola, A. A. Efros, et al., “The unreasonable effectiveness of deep features as a perceptual metric,” in Proc. IEEE/CVF Conf. Comput. Vis. Pattern Recognit. (CVPR), Salt Lake City, UT, USA, 2018, pp. 586–595.

[60] Y. Jiang, X. Gong, D. Liu, et al., “Enlightengan: Deep light enhancement without paired supervision,” IEEE Trans. Image Process., vol. 30, pp. 2340–2349, Jan. 2021.

[61] Y. Zhang, J. Zhang, and X. Guo, “Kindling the darkness: A practical low-light image enhancer,” in Proc. 27th ACM Int. Conf. Multimedia, Nice City, FR, 2019, pp. 1632–1640

[62] J. He, X. Zuo, J. Gao, *et al*., “Llformer: An efficient and real-time lidar lane detection method based on transformer,” in Proc. 5th Int. Conf. Pattern Recognition and Intelligent Systems, 2023, pp. 18–23.

[63] D. Ye, Z. Ni, W. Yang, *et al*., “Glow in the dark: Low-light image enhancement with external memory,” IEEE Trans. Multimedia, vol. 26, pp. 2148–2163, 2023.

[64] P. Singh and A. K. Bhandari, “A review on computational low-light image enhancement models: Challenges, benchmarks, and perspectives,” Arch. Comput. Methods Eng., vol. 32, no. 5, pp. 2853–2885, 2025.

[65] J. Cao, et al., “FIPNet: Self-supervised low-light image enhancement combining feature and illumination priors,” Neurocomputing, vol. 623, Art. no. 129426, 2025.

[66] A. Mittal, R. Soundararajan, and A. C. Bovik, “Making a ‘completely blind’ image quality analyzer,” IEEE Signal Process. Lett., vol. 20, no. 3, pp. 209–212, Mar. 2013.

[67] A. Mittal, A. K. Moorthy, and A. C. Bovik, “No-reference image quality assessment in the spatial domain,” IEEE Trans. Image Process., vol. 21, no. 12, pp. 4695–4708, Dec. 2012.

[68] Z. Ying, G. Li, Y. Ren, R. Wang, and W. Wang, “A fusion-based enhancing method for weak illumination images,” in Proc. IEEE Int. Conf. Image Processing (ICIP), Beijing, China, 2017, pp. 1–5.

[69] M. Yang and A. Sowmya, “An underwater color image quality evaluation metric,” IEEE Trans. Image Process., vol. 24, no. 12, pp. 6062–6071, Dec. 2015.